%% file: main.tex
\documentclass[runningheads]{llncs}
\usepackage{graphicx}
\usepackage{amsmath,amssymb} 
\usepackage{color}
\usepackage{mathtools,xparse}
\usepackage{xspace}
\usepackage{subcaption}
\usepackage[section]{placeins}
\usepackage{multirow}
\usepackage{breakcites}

\newcommand{\vect}[1]{{\bf {#1}}}

\newcommand{\norm}[1]{{||{#1}||}}

\makeatletter
\DeclareRobustCommand\onedot{\futurelet\@let@token\@onedot}
\def\@onedot{\ifx\@let@token.\else.\null\fi\xspace}

\def\eg{\emph{e.g}\onedot} 
\def\ie{\emph{i.e}\onedot}

\def\etal{\emph{et al}\onedot}
\makeatother

\begin{document}
\title{Learning Type-Aware Embeddings for Fashion Compatibility} 

\titlerunning{Learning Type-Aware Embeddings for Fashion Compatibility}

\authorrunning{Vasileva~\etal}

\author{Mariya I. Vasileva \and Bryan A. Plummer \and Krishna Dusad \and Shreya Rajpal \and Ranjitha Kumar \and David Forsyth}


\institute{Department of Computer Science,\\
	University of Illinois at Urbana-Champaign\\
	\email{ \{mvasile2,bplumme2,dusad2,srajpal2,ranjitha,daf\}@illnois.edu}
}
\maketitle              
\input{abstract}
\input{introduction}
\input{data}

\input{model}

\input{maryland_experiments}

\input{poly_experiments}

\input{conclusion}
\smallskip

\noindent\textbf{Acknowledgements:} This work is supported in part by ONR MURI Award N00014-16-1-2007, in part by NSF under Grant No. NSF IIS-1421521, and in part by a Google MURA Award and an Amazon Research Faculty Award.
%
%
%
%
%
\bibliographystyle{splncs04}
\bibliography{egbib}
\clearpage
\appendix
\input{supplementary}
\end{document}

%% file: abstract.tex
\begin{abstract}
Outfits in online fashion data are composed of items of many different types (\eg top, bottom, shoes) that share some stylistic relationship with one another. A representation for building outfits requires a method that can learn both notions of {\em similarity} (for example, when two tops are interchangeable) and {\em compatibility} (items of possibly different type that can go together in an outfit). This paper presents an approach to learning an image embedding that respects item type, and jointly learns notions of item similarity and compatibility in an end-to-end model. To evaluate the learned representation, we crawled 68,306 outfits created by users on the Polyvore website. Our approach obtains 3-5\% improvement over the state-of-the-art on outfit compatibility prediction and fill-in-the-blank tasks using our dataset, as well as an established smaller dataset, while supporting a variety of useful queries\footnote{Code and data:  \url{https://github.com/mvasil/fashion-compatibility}}.


\keywords{Fashion, embedding methods, appearance representations}
\end{abstract}

%% file: introduction.tex
\section{Introduction}

Outfit composition is a difficult problem to tackle due to the complex interplay of human creativity, style expertise, and self-expression involved in the process of
transforming a collection of seemingly disjoint items into a cohesive concept. Beyond selecting
which pair of jeans to wear on any given day, humans battle fashion-related problems ranging from, \emph{``How can
I achieve the same look as celebrity X on a vastly inferior budget?"}, to \emph{``How much would this scarf contribute
to the versatility of my personal wardrobe?"}, to \emph{``How should I dress to communicate motivation and competence
at a job interview?"}. This paper provides a step towards answering such diverse and logical queries.

\begin{figure*}[t!]
\centering
\begin{subfigure}[t]{0.23\textwidth}
\includegraphics[width=\textwidth]{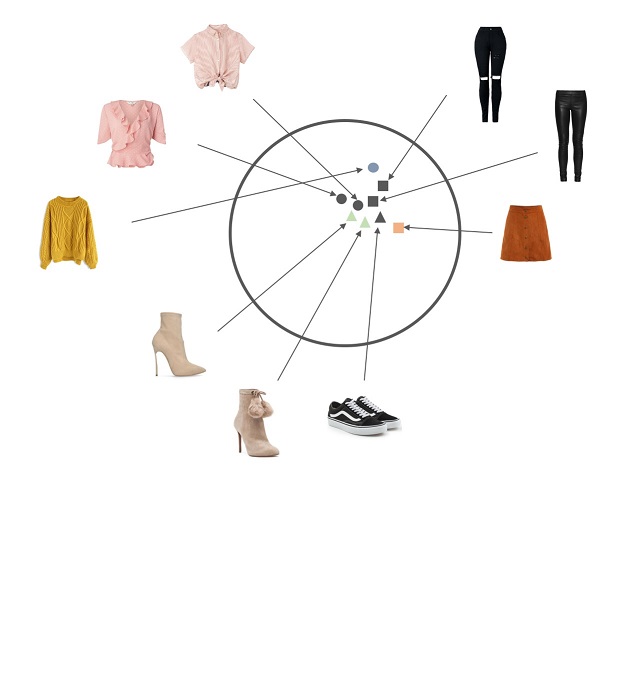}
\end{subfigure}
~ %
\begin{subfigure}[t]{0.63\textwidth}
\centering
\includegraphics[width=\textwidth]{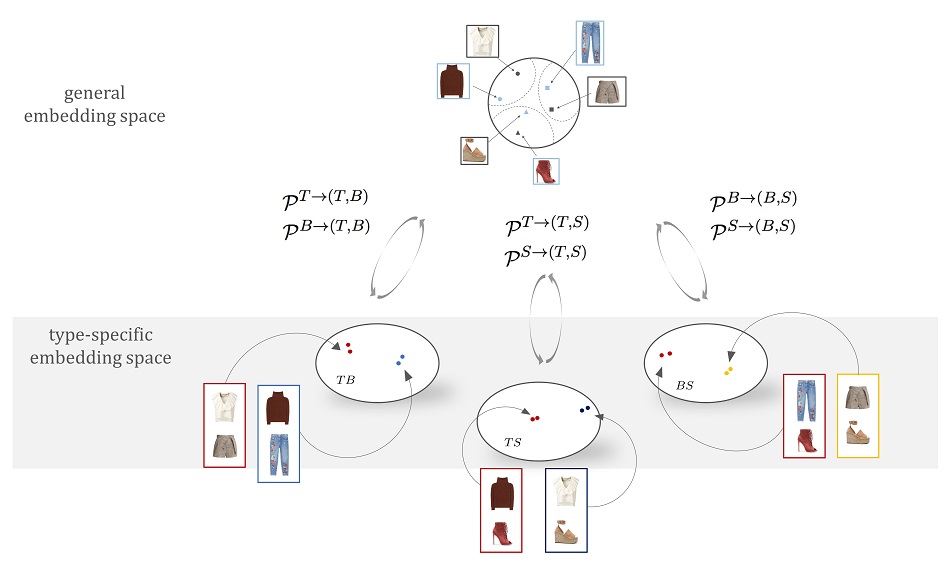}
\end{subfigure}
\caption{{\bf Left:} Conventional embedding strategies embed objects of all types in one underlying space. Objects that are compatible must lie close; as a result, all shoes that match a given top are \textbf{obliged} to be close.  {\bf Right:} Our type-respecting embedding, using ``top'', ``bottom'' and
  ``shoes'' as examples.  We first learn a single, shared embedding space.  Then, we project from that shared embedding to subspaces identified by type.  This means that all shoes that match a given top \textbf{must} be close in shoe-top space, but can be very different in the general embedding space. This enables us to search for two pairs of shoes that 1) match the same top, and 2) look very different from one another}
\label{fig:motivation}
\end{figure*}

To learn how to compose outfits the underlying representation must support both notions of {\em similarity} (\eg, when two tops are interchangeable) and notions of {\em compatibility} (items of possibly different type that can go together in an outfit). Current research handles both kinds of relationships with an embedding strategy: one trains a mapping, typically implemented as a convolutional neural network, that takes input items to an embedding space (\eg~\cite{hanACMMM2017,HePacMcA16,Veit2015}).  The training process tries to ensure that similar items are embedded nearby, and items that are different have widely separated embeddings (\ie the left side of Figure~\ref{fig:motivation}). 

These strategies, however, do not respect types (\eg shoes embed into the same space hats do), which has important consequences. Failure to respect types when training an embedding {\em compresses variation}: for instance, all shoes matching a particular hat are forced to embed close to one another, thus making them appear compatible even if they are not, which severely limits the model's ability to address diverse queries. Worse, this strategy encourages {\em improper triangles}: if a pair of shoes match a hat, and that hat in turn matches a blouse, then a natural consequence of models without type-respecting embeddings is that the shoes are forced to also match the blouse. This is because the shoes must embed close to the hat to match, the hat must embed close to the shoes to match, thus ensuring the shoes embed close to the blouse as well. Instead, they should be allowed to match in one context, and not match in another. An alternative way to describe the issue is that compatibility is \emph{not} naturally a transitive property, but \emph{being nearby} is.  Thus, an embedding that clusters items close together is not a natural way to measure compatibility without paying attention to context. By learning type-respecting spaces to measure compatibility, as in the right side of Figure~\ref{fig:motivation}, we avoid the
issues stemming from using a single embedding.

We begin by encoding each image in a general embedding space which we use to measure item similarity. The general embedding is trained using a visual-semantic loss between the image embedding, and features representing a text description of the corresponding item. This helps ensure that semantically similar items are projected in a nearby space.  In addition, we use a learned projection which maps our general embedding to a secondary embedding space that scores compatibility between two item types. We utilize a different projection for each pairwise compatibility comparison, unlike prior work which typically uses a single embedding space for compatibility scoring (\eg~\cite{hanACMMM2017,Veit2015}).  For example, if an outfit contains \emph{a hat}, \emph{a blouse}, and  \emph{a shoe}, we would learn projections for \emph{hat-blouse}, \emph{hat-shoe}, and \emph{blouse-shoe} embeddings.  The embeddings are trained along with a generalized distance metric, which we use to compute compatibility scores between items. Please refer to the appendix for a visualization of our approach.


Since many of the current fashion datasets either do not contain outfit compatibility annotations~\cite{Liu_2016_CVPR}, or are limited in size and the type of annotations they provide~\cite{hanACMMM2017}, we collect our own dataset which we describe in Section~\ref{sec:test_train_splits}. In Section~\ref{sec:type_embedding} we discuss our type-aware embedding model, which enables us to perform complex queries on our data.  Our experiments outlined in Section~\ref{sec:experiments} demonstrate the effectiveness of our approach, reporting a 4\% improvement in a fill-in-the-blank outfit completion experiment, and a 5\% improvement in an outfit compatibility prediction task over the prior state-of-the-art.

\section{Related Work}

Embedding methods provide the means to learn complicated relationships by simply providing samples of positive and negative pairs.  These approaches tend to be trained as a siamese network~\cite{Siamese} or using triplet losses~\cite{triplet} and have demonstrated impressive performance on challenging tasks like face verification~\cite{schroffCVPR2015}. This is attractive for many fashion-related tasks which typically require the learning of hard-to-define relationships between items (\eg ~\cite{Gomez_2017_CVPR,Kiapour_2015_ICCV,hanACMMM2017,HePacMcA16,Veit2015}). Veit~\etal~\cite{Veit2015} demonstrate successful similarity and compatibility matching for images of clothes on a large scale, but do not distinguish items by type (\eg top, shoes, scarves) in their embedding.  This is extended in Han~\etal~\cite{hanACMMM2017} by feeding the visual representation for each item in an outfit into an LSTM in order to jointly reason about the outfit as a whole.  There are several approaches which learn multiple embeddings (\eg~\cite{bell15productnet,Chen_2015_CVPR,Kiapour_2015_ICCV,Veit2017,Song_2017_ICCV_Workshops}), but tend to assume that instances of distinct types are separated (\ie comparing bags only to other bags), or use knowledge graph representation (\eg~\cite{xiaoAAAI2017}). Multi-modal embeddings appear to reveal novel feature structures (\eg~\cite{Gomez_2017_CVPR,hanACMMM2017,Rubio_2017_ICCV_Workshops,Salvador_2017_CVPR}), which we also take advantage of in Section~\ref{sec:constraints}.  
Training these type of embedding networks remains difficult because arbitrarily chosen triples can provide poor constraints~\cite{Wu_2017_ICCV,Zhuang_2016_CVPR}.
\smallskip

{\noindent}{\bf Fashion Studies.} Much of the recent fashion related work in the computer vision community has focused on tasks like product search and matching~\cite{Corbiere_2017_ICCV_Workshops,Garcia_2017_ICCV_Workshops,Kiapour_2015_ICCV,Liu_2016_CVPR,Zhao_2017_CVPR}, synthesizing garments from word descriptions~\cite{Zhu_2017_ICCV}, or using synthesis results as a search key~\cite{Zhao_2017_CVPR}.  Kiapour~\etal~\cite{HipsterWarsECCV14} trained an SVM over a set of hand-crafted features to categorize items into clothing styles. Vaccaro~\etal~\cite{Vaccaro:2016:EFS:2984511.2984573} trained a topic model over outfits in order to learn latent fashion concepts. Others have identified clothing styles using meta-data labels~\cite{Simo-Serra_2016_CVPR} or learning a topic model over a bag of visual attributes~\cite{Hsiao_2017_ICCV}.  Liu~\etal~\cite{Liu:2012:HMC:2393347.2393433} measure compatibility between items by reasoning about clothing attributes learned as latent variables in their SVM-based recommendation system.
Others have focused on attribute recognition~\cite{chenECCV2012,Di:2013:SFF:2514942.2515253,BMVC2015_51}, identifying relative strength of attributes between items (\ie a shoe is more/less pointy than another shoe)~\cite{singhECCV2016,yuCVPR2014,Yu_2015_ICCV,Yu_2017_ICCV}, or focused on predicting the popularity of clothing items~\cite{Al-Halah_2017_ICCV,Simo-Serra_2015_CVPR}.

%% file: data.tex
\section{Data}\label{sec:test_train_splits}

\begin{table}[t]
\centering
\setlength{\tabcolsep}{1pt}
\caption{Comparison in dataset statistics.  Our dataset's variants (last two rows) contains more outfits than related datasets along with detailed descriptions and fine-grained semantic categories}
\label{tab:data_stats_comparison}
\begin{tabular}{|l|c|c|c|c|c|}
\hline
\multirow{2}{*}{Dataset} & \multirow{2}{*}{\#Outfits} & \multirow{2}{*}{\#Items} & Max Items/ & \multirow{2}{*}{Text Available?} & Semantic\\
 & & & Outfit & & Category?\\
\hline
\hline
Maryland Polyvore~\cite{hanACMMM2017} & 21,889 & 164,379 & 8 & Titles Only & --\\
Polyvore Outfits-D & 32,140 & 175,485 & 16 & Titles \& Descriptions & \checkmark\\
Polyvore Outfits & 68,306 & 365,054 & 19 & Titles \& Descriptions & \checkmark\\
\hline
\end{tabular}
\end{table}

\noindent{\bf Polyvore Dataset:} The Polyvore fashion website enables users to create outfits as
compositions of clothing items, each containing rich multi-modal information such as product images, text descriptions,
associated tags, popularity score, and type information.  Han~\etal supplied a dataset of Polyvore outfits (referred to as the Maryland Polyvore dataset~\cite{hanACMMM2017}). This dataset is relatively small, does not contain item types or detailed text descriptions (see Table~\ref{tab:data_stats_comparison}), and has some inconsistencies in the test set that make
quantitative evaluation unreliable (additional details in Section~\ref{sec:experiments}). To resolve these issues, we collected our own dataset from Polyvore annotated with
outfit and item ID, fine-grained item type, title, text descriptions, and outfit images. Outfits that contain a single item or are missing type information are discarded, resulting in a total of 68,306
outfits and 365,054 items. Statistics comparing the two datasets are provided in Table~\ref{tab:data_stats_comparison}. 
\smallskip

\noindent{\bf Test-train splits:} Splits for outfit data are quite delicate, as one must consider whether a garment in the train set
should be allowed to appear in unseen test outfits, or not at all.  As some garments are ``friendly'' and appear in many outfits,
this choice has a significant effect on the dataset.  We provide two different versions of our dataset with respective
train and test splits. An ``easier" split contains 53,306 outfits for training, 10,000 for testing, and 5,000 for
validation, whereby no outfit appearing in one of the three sets is seen in the other two,
but it is possible that an item participating in a train outfit is also encountered in a test outfit. A more
``difficult" split is also provided whereby a graph segmentation algorithm was used to ensure that no garment appears in
more than one split. Each item is a node in the graph, and an edge connects two nodes if the corresponding items appear together in an outfit. This procedure requires discarding ``friendly'' garments, or else  the number of outfits collapses due to superconnectivity of the underlying graph. By
discarding the smallest number of nodes necessary, we end up with a total of 32,140 outfits and
175,485  items, 16,995 of which are used for training and 15,145 for testing and validation. \\

%% file: model.tex
\section{Respecting Type in Embedding}
\label{sec:type_embedding}

For the $i$'th data item $\vect{x}_i$, an embedding method uses some
regression procedure  (currently, a multilayer convolutional neural network) to compute a nonlinear feature embedding
$\vect{y}_i=\vect{f}(\vect{x}_i; \theta) \in \mathbb{R}^d$. The goal is to learn the parameters $\theta$ of the mapping
$\vect{f}$  such that for a pair of items $(\vect{x}_i, \vect{x}_j)$, the Euclidean distance between the embedding
vectors $\vect{y}_i$ and $\vect{y}_j$ reflects their compatibility. We would like to achieve a ``well-behaved" embedding
space in which  that distance is small for items that are labelled as compatible, and large for incompatible pairs. 

Assume we have a taxonomy of $\cal{T}$ types, and let us denote the type of an item as a superscript, such that $\vect{x}_i^{(\tau)}$ represents the $i$'th item of type $\tau$, where $\tau = 1, \ldots, \cal{T}$. A triplet is
defined as a set of images $ \{ \vect{x}_i^{(u)}, \vect{x}_j^{(v)}, \vect{x}_k^{(v)} \}$ with the following relationship: the anchor image $\vect{x}_i$ is of some type $u$, and both $\vect{x}_j$ and $\vect{x}_k$ are of a different type $v$. The pair $(\vect{x}_i,
\vect{x}_j)$ is compatible, meaning that the two items appear together in an outfit, while $\vect{x}_k$ is a randomly sampled
item of the same  type as $\vect{x}_j$ that has not been seen in an outfit with $\vect{x}_i$.  Let us write the standard triplet loss in the general form 

\begin{equation}
\ell (i, j, k) = \max \{ 0, d(i, j) - d(i, k) + \mu \} \;,
\label{eq:tripletloss}
\end{equation}

\noindent where $\mu$ is some margin.

We will denote by ${\cal M}^{(u, v)}$ the type-specific embedding space in which objects of types $u$ and $v$ are matched.  Associated with this space is a projection ${\cal
  P}^{u\rightarrow(u, v)}$ which maps the embedding of an object of type $u$ to ${\cal  M}^{(u, v)}$. 
Then,
for a pair of data items $(\vect{x}_i^{(u)}$, $\vect{x}_j^{(v)})$ that are compatible, we require the distance $\norm{{\cal
    P}^{u\rightarrow(u, v)}(\vect{f}(\vect{x}_i^{(u)};  \theta)) - {\cal P}^{v\rightarrow(u,
    v)}(\vect{f}(\vect{x}_j^{(v)}; \theta))}$ to be small.  This {\em does not} mean that the embedding vectors $\vect{f}(\vect{x}_i^{(u)};   \theta)$
and $\vect{f}(\vect{x}_j^{(v)};   \theta)$ for the two items in the general embedding space are similar - the differences just have to lie close to the kernel of ${\cal
  P}^{u\rightarrow(u, v)}$.   

This general form requires the learning of 2 $(d \times d)$ matrices per pair of types for a $d$-dimensional general
embedding. In this paper, we investigate two simplified versions: (a) assuming diagonal projection matrices such that ${\cal P}^{u\rightarrow(u, v)} = {\cal
  P}^{v\rightarrow(u, v)} =  \text{diag}(\vect{w}^{(u, v)})$, where $\vect{w}^{(u, v)}  \in \mathbb{R}^d$ is a vector of learned weights, and (b) the same case, but with $\vect{w}^{(u, v)}$ being a fixed binary vector chosen in advance for each pairwise type-specific space, and acting as a gating function that selects the relevant dimensions of the embedding most responsible for determining compatibility. Compatibility is then measured with
\begin{equation}
d_{ij}^{uv}=d(\vect{x}_i^{(u)}, \vect{x}_j^{(v)}, \vect{w}^{(u, v)}) = || \vect{f}(\vect{x}_i^{(u)}; \theta) \odot \vect{w}^{(u, v)} - \vect{f}(\vect{x}_j^{(v)}; \theta) \odot \vect{w}^{(u, v)} ||_2^2 \;,
\end{equation}
\noindent where $\odot$ denotes component-wise multiplication, and learned with the modified triplet loss:
\begin{equation}
\mathcal{L}_{\text{comp}} (\vect{x}_i^{(u)}, \vect{x}_j^{(v)}, \vect{x}_k^{(v)}, \vect{w}^{(u, v)}; \theta) = \max \{ 0, d_{ij}^{uv}-d_{ik}^{uv}+\mu\} \;,	
\label{eq:comp}
\end{equation}
\noindent where $\mu$ is some margin.

\subsection{Constraints on the learned embedding}
\label{sec:constraints}

To regularize the learned notion of compatibility, we further make use of the text descriptions accompanying each item image and feed them as input to a text embedding network. Let the embedding vector outputted by that network for the description $\vect{t}_i^{(u)}$ of image $\vect{x}_i^{(u)}$ be denoted $\vect{g}(\vect{t}_i^{(u)}; \phi)$, and substitute $\vect{g}$ for $\vect{f}$ in $\ell$ as required; the loss used to learn similarity is then 

\begin{equation}
\mathcal{L}_{\text{sim}} = \lambda_1\ell(\vect{x}_j^{(v)}, \vect{x}_k^{(v)}, \vect{x}_i^{(u)}) + \lambda_2\ell (\vect{t}_j^{(v)}, \vect{t}_k^{(v)}, \vect{t}_i^{(u)}) \;,
\label{eq:sim}
\end{equation}

\noindent where $\lambda_{1-2}$ are scalar parameters.

We also train a visual-semantic embedding in the style of Han~\etal~\cite{hanACMMM2017} by requiring that image $\vect{x}_i^{(u)}$ is embedded closer to its description $\vect{t}_i^{(u)}$ in visual-semantic space than the descriptions of the other two images in a triplet:

\begin{equation}
\mathcal{L_\text{vse}}_i = \ell(\vect{x}_i^{(u)}, \vect{t}_i^{(u)}, \vect{t}_j^{(v)}) + \ell(\vect{x}_i^{(u)}, \vect{t}_i^{(u)}, \vect{t}_k^{(v)})
\label{eq:vse}
\end{equation}

\noindent and imposing analogical constraints on $\vect{x}_j^{(v)}$ and $\vect{x}_k^{(v)}$.

To encourage sparsity in the learned weights $\vect{w}$ so that we achieve better disentanglement of the embedding dimensions contributing to pairwise type compatibility, we add an $\ell_1$ penalty on the projection matrices $ \mathcal{P}^{\cdot \rightarrow(\cdot, \cdot)}$. We further use $\ell_2$ regularization on the learned image embedding $\vect{f}(\mathbf{x}; \theta)$. The final training loss therefore becomes:
	
\begin{equation}
\mathcal{L} (\mathbf{X}, \mathbf{T}, \mathcal{P}^{\cdot \rightarrow(\cdot, \cdot)}, \vect{\lambda}, \theta, \phi) = \mathcal{L}_\text{comp} + \mathcal{L}_{\text{sim}} + \lambda_3 \mathcal{L}_{\text{vse}} + \lambda_4 \mathcal{L}_{\ell_2} + \lambda_5 \mathcal{L}_{\ell_1} \\
\label{eq:full_loss}
\end{equation}
	
\noindent where $\mathbf{X}$ and $\mathbf{T}$ denote the image embeddings and corresponding text embeddings in a batch, $\mathcal{L}_{\text{vse}} = \mathcal{L}_{\text{vse}_i} + \mathcal{L}_{\text{vse}_j} + \mathcal{L}_{\text{vse}_k}$, and $\lambda_{3-5}$ are scalar parameters.  We preserve the dependence on $ \mathcal{P}^{\cdot
  \rightarrow(\cdot, \cdot)}$ in notation to emphasize the type dependence of our embedding. As
Section~\ref{sec:experiments} shows, this term has significant effects.

%% file: maryland_experiments.tex
\section{Experiment Details}
\label{sec:experiments}

Following Han~\etal~\cite{hanACMMM2017} we evaluate how well our approach performs on two tasks.  In the \emph{fashion compatibility} task, a candidate outfit is scored as to whether its constitute items are compatible with each other.  Performance is evaluated using the average under a receiver operating characteristic curve (AUC).  The second task is to select from a set of candidate items (four in this case) in a fill-in-the-blank (FITB) fashion recommendation experiment.  The goal is to select the most compatible item with the remainder of the outfit, and performance is evaluated by accuracy on the answered questions.
\smallskip

\noindent{\bf Datasets.} For experiments on the Maryland Polyvore dataset~\cite{hanACMMM2017} in Section~\ref{sec:maryland_experiments} we use the provided splits which separate the outfits into 17,316 for training, 3,076 for testing, and 1,407 for validation.  For experiments using our dataset in Section~\ref{sec:polyvore_experiments} we use the different version splits described in Section~\ref{sec:test_train_splits}. We shall refer to the ``easier'' split as \emph{Polyvore Outfits}, and the split containing only disjoint outfits down to the item level as \emph{Polyvore Outfits-D}.

\noindent{\bf Implementation.}  We use a 18-layer Deep Residual Network~\cite{He_2016_CVPR} which was pretrained on ImageNet~\cite{deng2009imagenet} for our image embedder with a general embedding size of $64$ dimensions unless otherwise noted.  Our model is trained with a learning rate of $5e^{-5}$, batch size of $256$, and a margin of $0.2$.  For our text representation, we use the HGLMM Fisher vector encoding~\cite{klein2014fisher} of word2vec~\cite{mikolov2013efficient} after having been PCA reduced down to $6000$ dimensions.  We set $\lambda_3=5e^{-1}$ from Eq.~(\ref{eq:full_loss}) for the Maryland Polyvore dataset ($\lambda_3=5e^{-5}$ for experiments on our dataset), and all other $\lambda$ parameters from Eq.~(\ref{eq:sim}) and Eq.~(\ref{eq:full_loss}) to $5e^{-4}$.
\smallskip

\noindent{\bf Sampling Testing Negatives.} In the test set provided by Han~\etal~\cite{hanACMMM2017}, a negative outfit for the compatibility experiment could end up containing only tops and no other items, and a fill-in-the blank question could have an earring be among the possible answers when trying to select a replacement for a shoe. This is a result of sampling negatives at random without restriction, and many of these negatives could simply be dismissed without considering item compatibility. Thus, to correct this issue and focus on outfits that cannot be filtered out in such a manner, we take into account item category when sampling negatives. In the compatibility experiments, we replace each item in a ground truth outfit by randomly selecting another item of the same category. For the fill-in-the-blank experiments, our incorrect answers are randomly selected items from the same category as the correct answer.
\smallskip

\noindent{\bf Comparative Evaluation.} In addition to performance of the state-of-the-art methods reported in prior work, we compare the following approaches:

\begin{itemize}
\item {\bf SiameseNet (ours).} The approach of Veit~\etal~\cite{Veit2015} which uses the same ResNet and general embedding size as used for our type-specific embeddings.
\item {\bf CSN, T1:1.} Learns a pairwise type-dependent transformation using the approach of Veit~\etal~\cite{Veit2017} to project a general embedding to a type-specific space which measures compatibility between two item categories.
\item {\bf CSN, T4:1.} Same as the previous approach, but where each learned pairwise type-dependent transformation is responsible for four pairwise comparisons (instead of one) which are assigned at random.  For example, a single projection may be used to measure compatibility in the (shoe-top, bottom-hat, earrings-top, outwear-bottom) type-specific spaces. This approach allows us to assess the importance of having distinct learned compatibility spaces for each pair of item categories versus forcing the compatibility spaces to ``share'' multiple pairwise comparisons, thus allowing for better scalability as we add more fine-grained item categories to the model.
\item {\bf VSE.}  Indicates that a visual-semantic embedding as described in Section~\ref{sec:constraints} is learned jointly with the compatibility embedding.
\item {\bf Sim.}  Along with training the model to learn a visual-semantic embedding for compatibility between different categories of items as done with the VSE, the same embeddings are also used to measure similarity between items of the same category as described in Section~\ref{sec:constraints}.
\item {\bf Metric.}  In the triplet loss, rather than minimizing Euclidean distance between compatible items and maximizing the same for incompatible ones, an empirically more robust way is to optimize over the inner products instead. 
To generalize the distance metric, we take an element-wise product of the embedding vectors in the type-specific spaces and feed it into a fully-connected layer, the learned weights of which act as a generalized distance function. 
\end{itemize}

\begin{table}[t]
\centering
\caption{Comparison of different methods on the Maryland Polyvore dataset~\cite{hanACMMM2017} using their unrestricted randomly sampled negatives on the fill-in-the-blank and outfit compatibility tasks.  ``All Negatives'' refers to using their entire test split as is, while ``Composition Filtering'' refers to removing easily identifiable negative samples.  The numbers in {\bf (a)} are the results reported from Han~\etal~\cite{hanACMMM2017} or run using their code, and {\bf (b)} reports our results}
\label{tab:maryland_exps}
\begin{tabular}{|rl|c|c|c|c|}
\hline
& & \multicolumn{2}{|c|}{All Negatives} & \multicolumn{2}{|c|}{w/Composition Filtering}\\
\hline
& \multirow{2}{*}{Method} & FITB & Compat. & FITB & Compat. \\
& & Accuracy & AUC & Accuracy &  AUC\\
\hline
\hline
{\bf (a)} & SetRNN~\cite{LiCZL17SetRNN} & 29.6 & 0.53 & -- & --\\
& SiameseNet~\cite{Veit2015} & 52.0 & 0.85 & -- & --\\
& Bi-LSTM (512-D)~\cite{hanACMMM2017} & 66.7 & 0.89 & -- & --\\
& Bi-LSTM + VSE (512-D)~\cite{hanACMMM2017} & 68.6 & 0.90 & \textbf{81.5} & 0.78\\
\hline
{\bf (b)} & SiameseNet (ours) & 54.2 & 0.85 & 72.3 & 0.81\\
& CSN, T1:1 & 51.6 & 0.83 & 74.9 & 0.83\\
& CSN, T1:1 + VSE & 52.4 & 0.83 & 73.1 & 0.83\\
& CSN, T1:1 + VSE + Sim & 51.5 & 0.82 & 75.1 & 0.79\\
& CSN, T4:1 + VSE + Sim + Metric & 84.2 & 0.90 & 75.7 & \textbf{0.84}\\
& CSN, T1:1 + VSE + Sim + Metric & \textbf{86.1} & \textbf{0.98} & 78.6 & \textbf{0.84}\\
\hline
\end{tabular}
\end{table}

\begin{table}[t]
\centering
\caption{Comparison of different methods on the Maryland Polyvore Dataset~\cite{hanACMMM2017} on the fill-in-the-blank and outfit compatibility tasks using our category-aware negative sampling method.  {\bf (a)} contains the results of prior work using their code unless otherwise noted, and {\bf (b)} contains results using our approach}
\label{tab:maryland_exps_ours}
\begin{tabular}{|rl|c|c|}
\hline
&\multirow{2}{*}{Method} & FITB & Compat.\\
&& Accuracy & AUC\\
\hline
\hline
{\bf (a)}&Bi-LSTM + VSE (512-D)~\cite{hanACMMM2017} & 64.9 & \textbf{0.94}\\
&SiameseNet (ours) & 54.4 & 0.85\\
\hline
{\bf (b)}&CSN, T1:1 & 57.9 & 0.87\\
&CSN, T1:1 + VSE & 58.1 & 0.88 \\
&CSN, T1:1 + VSE + Sim & 59.0 & 0.87 \\
&CSN, T4:1 + VSE + Sim + Metric & 59.9 & 0.90\\
&CSN, T1:1 + VSE + Sim + Metric & 61.0 & 0.90\\
&CSN, T1:1 + VSE + Sim + Metric (512-D) & \textbf{65.0} & 0.93\\
\hline
\end{tabular}
\end{table}

\begin{table}[t]
\centering
\caption{Effect the embedding size has on performance on the fill-in-the-blank and the outfit compatibility tasks on the Maryland Polyvore dataset~\cite{hanACMMM2017} using our negative samples}
\label{tab:maryland_exps_dim}
\begin{tabular}{|rl|c|c|}
\hline
& \multirow{2}{*}{Method} & FITB & Compat.\\
& & Accuracy & AUC\\
\hline
\hline
& CSN, T1:1 + VSE + Sim + Metric (32-D) & 55.7 & 0.88\\
& CSN, T1:1 + VSE + Sim + Metric (64-D) & 61.0 & 0.90\\
& CSN, T1:1 + VSE + Sim + Metric (128-D) & 62.4 & 0.92\\
& CSN, T1:1 + VSE + Sim + Metric (256-D) & 62.8 & 0.92\\
& CSN, T1:1 + VSE + Sim + Metric (512-D) & \textbf{65.0} & \textbf{0.93}\\
\hline
\end{tabular}
\end{table}

\begin{figure}[t]
\centerline{\includegraphics[width=0.5\textwidth]{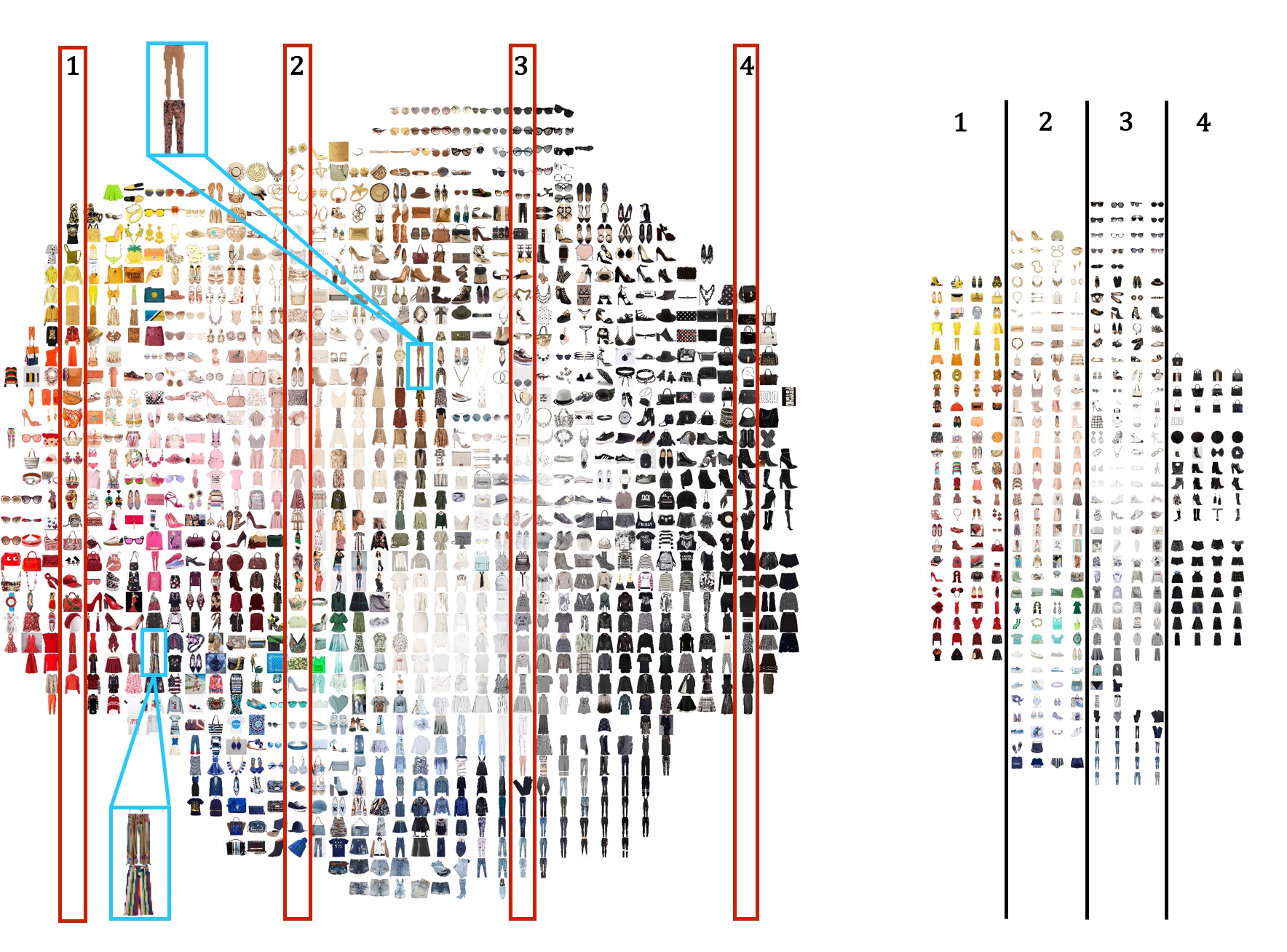}}
\caption{\textbf{Left:} t-SNE of the learned general embedding space on Polyvore Outfits. We see the learned embedding respects color variations and for types where shape is a unique identifier (\eg, pants and sunglasses) items are more closely grouped together. \textbf{Right:} Overlapping items for each cell of the highlighted four columns in the t-SNE plot. Note that each row contains items that are very similar to each other, which suggests a well-behaved embedding.
Best viewed in color at high resolution}
\label{tsne-big}
\end{figure}

\begin{figure*}
\centering
\begin{subfigure}[t]{0.25\textwidth}
\includegraphics[width=\textwidth]{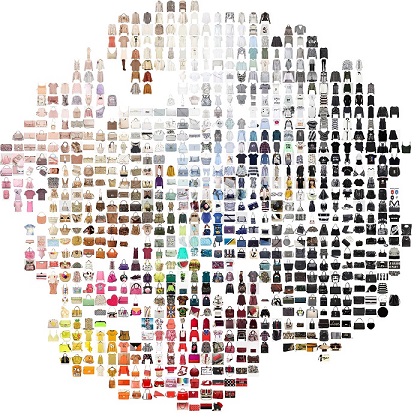}
\caption{}
\end{subfigure}
~%
\begin{subfigure}[t]{0.25\textwidth}
\centering
\includegraphics[width=\textwidth]{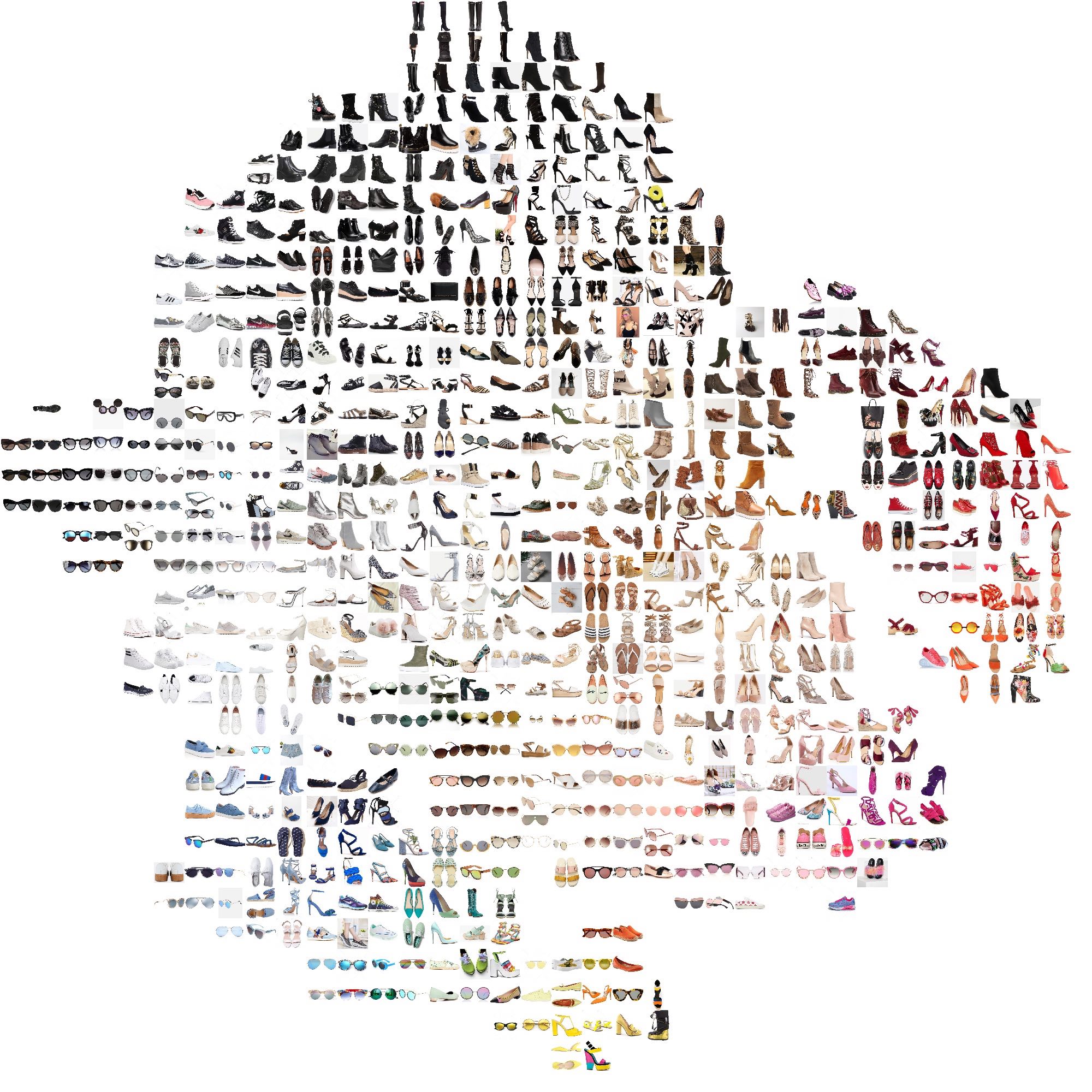}
\caption{}
\end{subfigure}
~%
\begin{subfigure}[t]{0.25\textwidth}
\centering
\includegraphics[width=\textwidth]{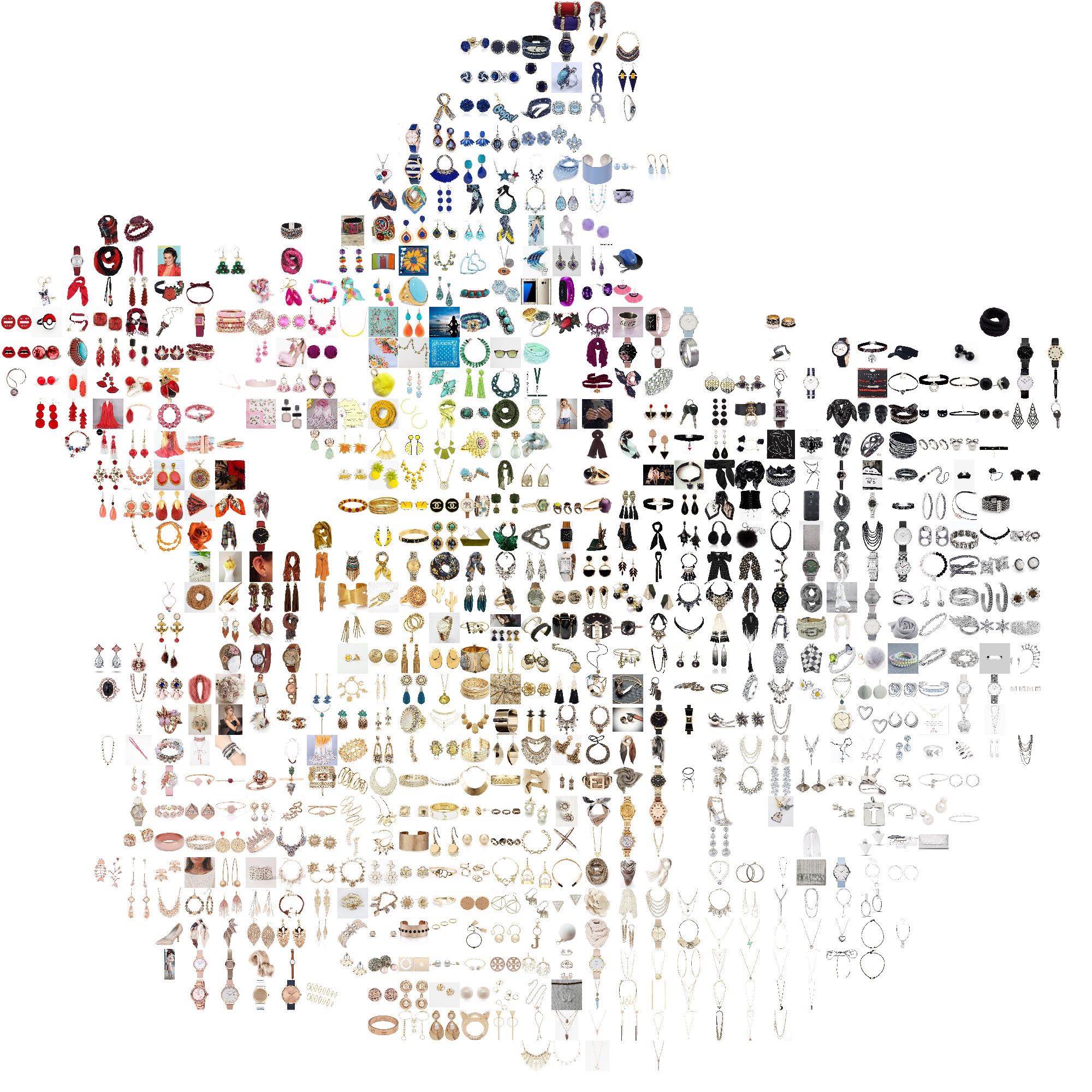}
\caption{}
\end{subfigure}
\caption{t-SNE of the learned type-specific embedding space on our Polyvore dataset for: \textbf{(a)} tops and bags; \textbf{(b)} shoes and sunglasses; \textbf{(c)} scarves and jewelery. As hypothesized, respecting type allows the embedding to specialize to features that dominate compatibility relationships for each pair of types: for example, color seems to matter more in (a) than in (c), where shape is an equally important feature, with a concentration of long pendants in the lower right and smaller pieces towards the top. Best viewed in color at high resolution}
\label{tsne-local}
\end{figure*}

\begin{figure}
\centering
\includegraphics[width=0.85\textwidth]{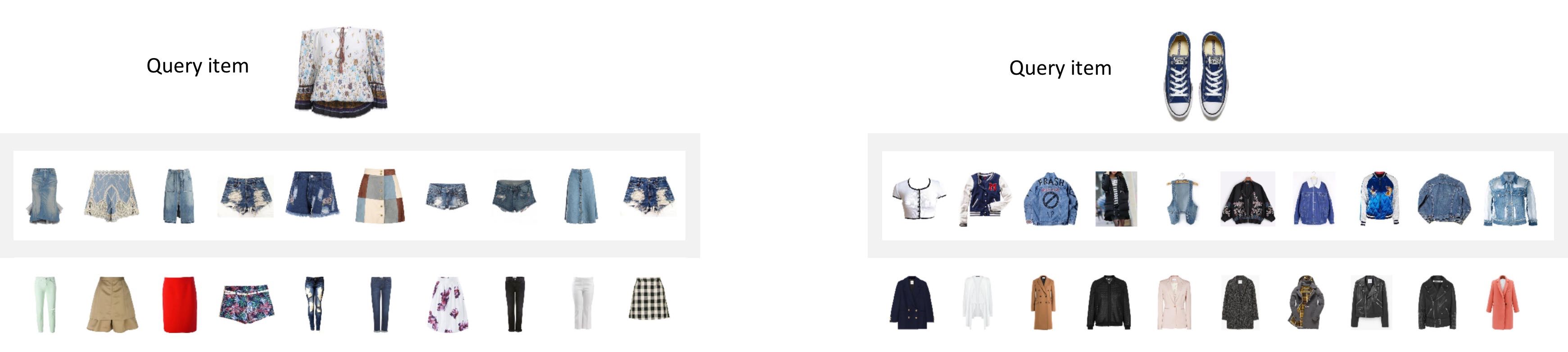}
\caption{Examples of learned compatibility relationships by our model. The query item (top row) is randomly pulled from some type $u$. Rows shaded in gray show our model's suggestions for items compatible with the query item of a randomly selected type $v$. Bottom row shows items of type $v$ sampled at random}
\label{fig:compat}
\end{figure}

\begin{figure}
\centering
\includegraphics[width=0.65\textwidth]{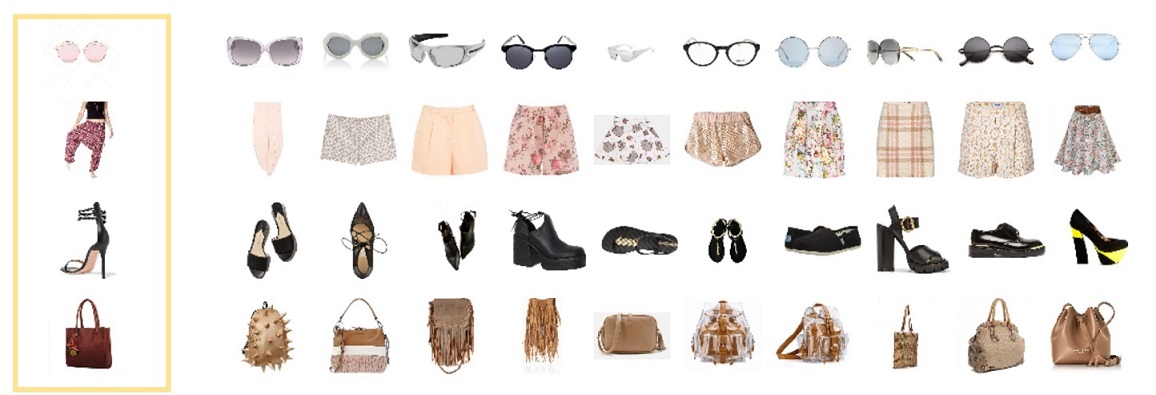}
\caption{Examples of learned similarity relationships by our model. Column outlined in yellow shows the query image for each row. Rows contain similar items of the same type as the query item. Note that the learned notion of similarity is more abstract than matching simply based on color or shape. For example, in the second row, the model is perfectly happy suggesting as alternatives both short- and long-leg pants as well as skirts, so long as the general light colors, flowery patterns and flowy materials are present. Similarly, in the third row, all similar shoes are black but they vastly differ in style (e.g., platform vs. sandal vs. bootie vs. loafer) and all have a unique statement element just like the straps detail of the query item: laces, a golden clasp, yellow detail, metal bridge
\label{fig:similarity}}
\end{figure}

\subsection{Maryland Polyvore}
\label{sec:maryland_experiments}
We report performance using the test splits of Han~\etal~\cite{hanACMMM2017} in Table~\ref{tab:maryland_exps} where the negative samples were sampled completely at random without restriction.  The first line of Table~\ref{tab:maryland_exps}(b) contains our replication of the approach of Veit~\etal~\cite{Veit2015}, using the same convolutional network as implemented in our models for a fair comparison.  We see on the second line of Table~\ref{tab:maryland_exps}(b) that performance on both tasks using all the negative samples in the test split is actually reduced, while after removing easily identifiable negative samples it is increased.  This is likely due to how the negatives were sampled.  We only learn type specific embeddings to compare the compositions of items which occur during training.  Thus, at test time, if we are asked to compare two tops, but no outfit seen during training contained two tops, we did not learn a type-specific embedding for this case and are forced to compare them using our general embedding.  This also explains why our performance drops using the negative samples of Han~\etal when we include the similarity constraint on the third line of Table~\ref{tab:maryland_exps}(b), since we explicitly try to learn similarity in our general embedding rather than compatibility.  The same effect also accounts for the discrepancy when we train our learned metric shown in the last two lines of Table~\ref{tab:maryland_exps}(b).  Although we report much better performance than prior work using all the negative samples, our performance is much closer to the LSTM-based method of Han~\etal~\cite{hanACMMM2017} after removing the easy negatives.

Since many of the negative samples of Han~\etal~\cite{hanACMMM2017} can be easily filtered out, and thus make it difficult to evaluate our method due to their invalid outfit compositions, we report performance on the fill-in-the-blank and outfit compatibility tasks where the negatives are selected by replacing items of the same category in Table~\ref{tab:maryland_exps_ours}.  The first line of Table~\ref{tab:maryland_exps}(b) shows that using our type-specific embeddings, we obtain a 2-3\% improvement over learning a single embedding to compare all types.  In the second and third lines of Table~\ref{tab:maryland_exps_ours}(b), we see that including our visual semantic embedding, along with training our general embedding to explicitly learn similarity between objects of the same category, provides small improvements over simply learning our type-specific embeddings.  We also see a pronounced improvement using our learned metric, resulting in a 3-4\% improvement on both tasks over learning just the type-specific embeddings.  The last line of Table~\ref{tab:maryland_exps_ours}(b) reports the results of our approach using the same embedding size as Han~\etal~\cite{hanACMMM2017}, showing that we obtain similar performance.  This is particularly noteworthy since Han~\etal uses a more powerful feature representation (Inception-v3~\cite{Szegedy_2016_CVPR} vs.\ ResNet-18), and takes into account the entire outfit when making comparisons, both of which would likely further improve our model.  A full accounting of how the dimensions of the final embedding affects the performance of our approach is provided in Table~\ref{tab:maryland_exps_dim}.

%% file: poly_experiments.tex
\begin{table}[t]
\centering
\caption{Comparison of different methods on the two versions of our dataset on the fill-in-the-blank and outfit compatibility tasks using our category-aware negative sampling method.  {\bf (a)} contains the results of prior work using their code unless otherwise noted, and {\bf (b)} contains results using our approach}
\label{tab:outfits_exps}
\begin{tabular}{|rl|c|c|c|c|}
\hline
&& \multicolumn{2}{|c|}{Polyvore Outfits-D} & \multicolumn{2}{|c|}{Polyvore Outfits}\\
\hline
&\multirow{2}{*}{Method} & FITB & Compat. & FITB & Compat. \\
& & Accuracy & AUC & Accuracy &  AUC\\
\hline
\hline
{\bf (a)}&Bi-LSTM + VSE (512-D)~\cite{hanACMMM2017} & 39.4 & 0.62 & 39.7 & 0.65\\
&SiameseNet (ours) & 51.8 & 0.81 & 52.9 & 0.81\\
\hline
{\bf (b)}&CSN, T1:1 & 52.5 & 0.82 & 54.0 & 0.83\\
&CSN, T1:1 + VSE & 53.0 & 0.82 & 54.5 & 0.84\\
&CSN, T1:1 + VSE + Sim & 53.4 & 0.82 & 54.7 & 0.85\\
&CSN, T4:1 + VSE + Sim + Metric  & 53.7 & 0.82 & 55.1 & 0.85\\
&CSN, T1:1 + VSE + Sim + Metric & 54.1 & 0.82 & 55.3 & \textbf{0.86}\\
&CSN, T1:1 + VSE + Sim + Metric (512-D) & \textbf{55.2} & \textbf{0.84} & \textbf{56.2} & \textbf{0.86}\\
\hline
\end{tabular}
\end{table}

\begin{table}[t]
\centering
\caption{Effect the embedding size has on performance on the fill-in-the-blank and the outfit compatibility tasks using the two versions of our dataset}
\label{tab:outfits_exps_dim}
\begin{tabular}{|rl|c|c|c|c|}
\hline
& & \multicolumn{2}{|c|}{Polyvore Outfits-D} & \multicolumn{2}{|c|}{Polyvore Outfits}\\
\hline
& \multirow{2}{*}{Method} & FITB & Compat. & FITB & Compat. \\
& & Accuracy & AUC & Accuracy &  AUC\\
\hline
\hline
& CSN, T1:1 + VSE + Sim + Metric (32-D) & 53.2 & 0.81 & 53.9 & 0.85\\
& CSN, T1:1 + VSE + Sim + Metric (64-D) & 54.1 & 0.82 & 55.3 & \textbf{0.86}\\
& CSN, T1:1 + VSE + Sim + Metric (128-D) & 54.3 & 0.83\ & 55.2 & \textbf{0.86}\\
& CSN, T1:1 + VSE + Sim + Metric (256-D) & 54.8 & \textbf{0.84} & 55.6 & \textbf{0.86}\\
& CSN, T1:1 + VSE + Sim + Metric (512-D) & \textbf{55.2} & \textbf{0.84} & \textbf{56.2} & \textbf{0.86}\\
\hline
\end{tabular}
\end{table}

\begin{figure}[t]
\centering
\centerline{\includegraphics[width=.9\textwidth]{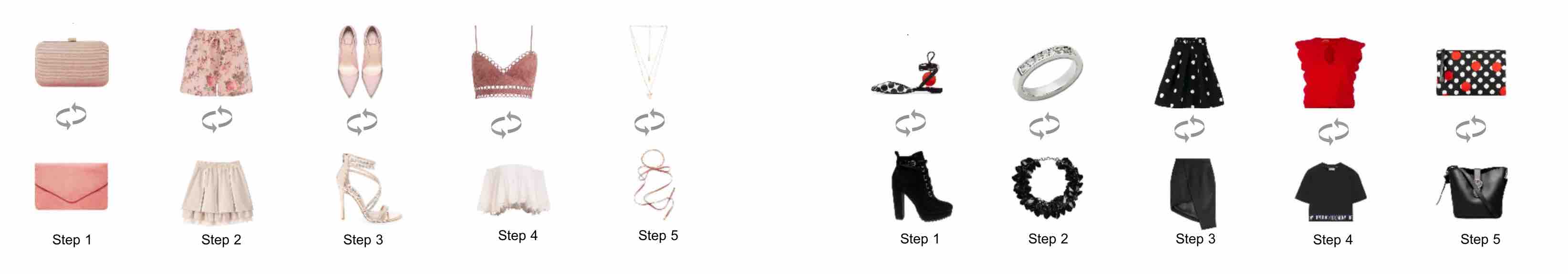}}
\caption{Examples of outfit generation by recursive item swaps. The top row represents a valid (\ie, human-curated) outfit. At each step, we replace an item from the starting outfit with one that is of the same type and equally compatible with the rest of the outfit, but different from the removed item. For full figure, refer to the appendix.
Best viewed in color
\label{fig:swaps}}
\end{figure}

\begin{figure}
\centering
\centerline{\includegraphics[width=0.8\textwidth]{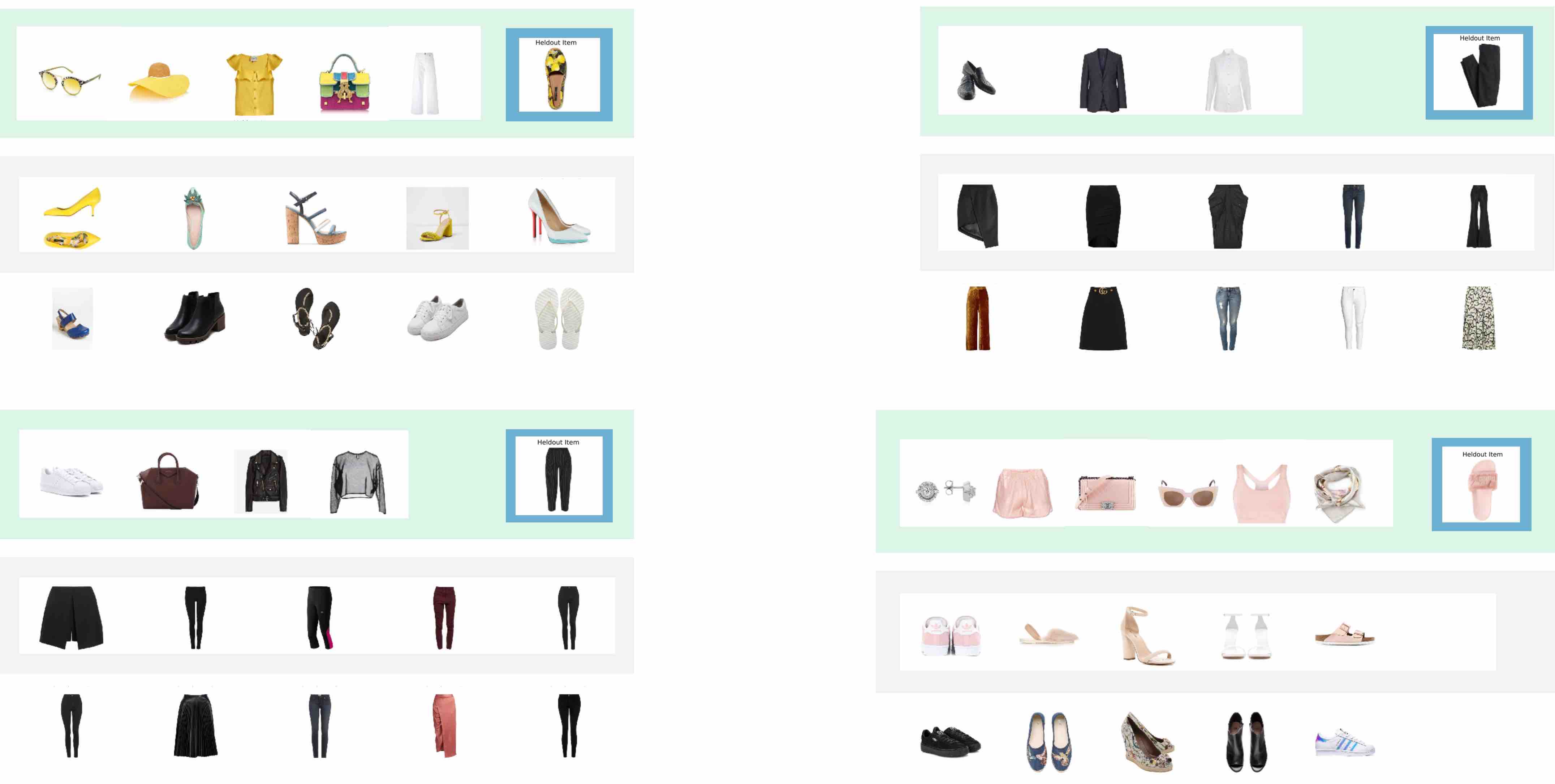}}
\caption{Examples of item swaps and outfit diversification. Row shaded in green represents a human-curated outfit. Highlighted in blue is a randomly selected heldout item from the outfit to be replaced. Rows shaded in gray displays alternatives that are all equally compatible with the rest of the outfit as the heldout item. The bottom row shows a random selection of alternatives of the same type as the heldout item. The suggested alternatives made by our model in the middle row, although equally compatible with the rest of the items in the outfit, are not forced to be similar to each other but differ vastly in color, style, shape and fine-grained type.
Best viewed in color
\label{fig:comp_random}}
\end{figure}

\subsection{Polyvore Outfits}
\label{sec:polyvore_experiments}

We report our results on the fill-in-the-blank and outfit compatibility experiments using our own dataset in Table~\ref{tab:outfits_exps}.  The first line of Table~\ref{tab:outfits_exps}(b) shows that learning our type specific embeddings gives a consistent improvement over training a single embedding to make all comparisons.  We note that our relative performance using the entire dataset is higher than our disjoint set, which we attribute to likely being due to the additional training data for learning each type-specific embedding.  Analogous to the Maryland dataset, the next three lines of Table~\ref{tab:outfits_exps} show a consistent performance improvement as we add in the remaining pieces of our model.

Interestingly, the two splits of our data obtain similar performance, with the better results using the easy version of our dataset only a little better than on the version where all items in all outfits are novel.  This suggests that having unseen outfits in the test set is more important than ensuring there are no shared items between the training and testing splits, and hence in reproductions of our experiments, using the larger version of our dataset is a fair approach.

\subsubsection{Why does respecting type help?} We visualize the global embedding space and some type-specific embedding spaces with t-SNE~\cite{vandermatten2008JMLR}.  Figure~\ref{tsne-big} shows the global embedding space; Figure~\ref{tsne-local} shows three distinct type-specific embedding spaces.  Note how the global space is strongly oriented toward color matches (large areas allocated to each range of color), but for example the scarf-jewelery space in Figure~\ref{tsne-local}(c) is not particularly focused on color representation, preferring to encode shape (long pendants vs. smaller pieces).  As a result, local type-specific spaces can specialize in different aspects of appearance, and so force the global space to represent all aspects fairly evenly.

\subsubsection{Geometric Queries.}

In light of the problems pointed out in the introduction, we show that our type-respecting embedding is able to handle the following geometric queries which previous models are unable or ill-equipped to perform. SiameseNet~\cite{Veit2015} is not able to answer such queries by construction, and it is not straightforward how the approach of Han~\etal~\cite{hanACMMM2017} would have to be repurposed in order to handle them. Our model is the first to demonstrate that this type of desirable query can be successfully addressed.

\begin{itemize}
\item Given an item $\vect{x}_i^{(u)}$ of a certain type, show a collection of items $\{ \vect{x}_j^{(v)} \}_{j = 1}^N$ of a different type that are all compatible with $\vect{x}_i^{(u)}$ but dissimilar from each other (see Figure~\ref{fig:compat}).
\item Given an item $\vect{x}_i^{(u)}$ of a certain type, show a collection of items $\{ \vect{x}_j^{(u)} \}_{j = 1}^N$ of the same type that are all interchangeable with $\vect{x}_i^{(u)}$ but have diverse appearance (see Figure~\ref{fig:similarity})
\item Given a valid outfit $\mathcal{S} = \{ \vect{x}_k^{(\tau)} \}_{k = 1, \tau = 1}^{K, \mathcal{T}}$, replace each item $\vect{x}_k^{(\tau)}$ in turn with an item $\tilde{\vect{x}}^{(\tau)}$ of the same type which is different from $\vect{x}_k^{(\tau)}$, but compatible with the rest of the outfit $\mathcal{S}_{\backslash \{ \vect{x}_k^{(\tau)} \}}$ (see Figure~\ref{fig:swaps}).
\item Given an item $\vect{x}_i^{(u)}$ from a valid outfit $\mathcal{S} = \{ \vect{x}_k^{(\tau)} \}_{k = 1, \tau = 1}^{K, \mathcal{T}}$, show a collection of \emph{replacement} items $\{ \vect{x}_j^{(u)} \}_{j = 1}^N$ of the same type that are all compatible with the rest of the outfit $\mathcal{S}_{\backslash \{ \vect{x}_i^{(u)} \}}$ but visually different from $\vect{x}_i^{(u)}$ (see Figure~\ref{fig:comp_random}).
\end{itemize}

%% file: conclusion.tex
\section{Conclusion}
Our qualitative and quantitative results show that respecting type in embedding methods produces several strong and useful effects.  First, on an established dataset, respecting type produces better performance at established tasks.  Second, on a novel, and richer, dataset, respecting type produces strong performance improvements on established tasks over previous methods.  Finally, an embedding method that respects type can represent both {\em similarity} relationships (whereby garments are interchangeable - say, two white blouses) and {\em compatibility} relationships (whereby a garment can be combined with another to form a coherent outfit). Visualizing the learned embedding spaces suggests that the reason we obtain significant improvements on the fill-in-the-blank and outfit compatibility tasks over prior state-of-the-art is that different type-specific spaces specialize in encoding different kinds of appearance variation. The resulting representation admits new and useful queries for clothing.  One can search for item replacements; one can find a set of possible items to complete an outfit that has high variation; and one can swap items in garments to produce novel outfits.  

%% file: supplementary.tex
\section{Model Visualization}

\begin{figure}
\centering
\includegraphics[width=\textwidth]{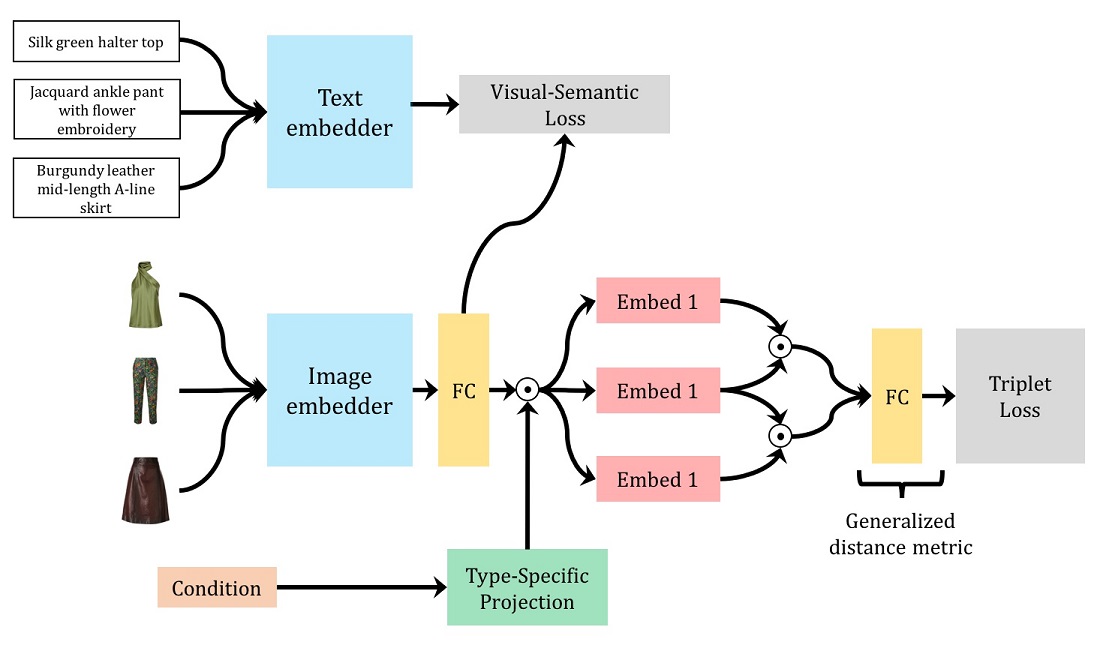}
\caption{An overview of model architecture. We use the type-specific projections to transform the general image embedding into pairwise compatibility spaces, along with a generalized distance metric to compare triplets. In addition to scoring type-dependent compatibility, we also train a visual semantic embedding using features of text descriptions accompanying each item, which regularizes the general embedding and the learned compatibility and similarity relationships}
\label{fig:model_diagram}
\end{figure}

\section{Ablation study}

In Table~\ref{tab:outfits_exps} we provide an ablation study to supplement the results in the paper.  Two additional settings are considered: \textit{FC} which uses a fully connected layer for its type-specific projection rather than a learned diagonal projection and \textit{Cosine} which uses cosine distance to train our model rather than euclidean distance.  Here we see that using a \textit{FC} layer provides a small performance improvement at a higher computation cost while our learned metric performs slightly better than using cosine distance.

\begin{table}[t]
\centering
\caption{Additional results for Table 5 from the paper.  {\bf (a)} contains results using our approach where some compatibility spaces are shared, {\bf (b)} contains results where we use a fully connected layer to project our general embedding into our compatibility space, {\bf (c)} compares our learned metric with using cosine distance, and {\bf (d)} contains additional ablations of our approach's components}
\label{tab:outfits_exps}
\begin{tabular}{|rl|c|c|c|c|}
\hline
&& \multicolumn{2}{|c|}{Polyvore Outfits-D} & \multicolumn{2}{|c|}{Polyvore Outfits}\\
\hline
&\multirow{2}{*}{Method} & FITB & Compat. & FITB & Compat. \\
& & Accuracy & AUC & Accuracy &  AUC\\
\hline
\hline
{\bf (a)}&CSN, T4:1 & 52.3 & 0.80 & 54.1 & 0.83\\
&CSN, T4:1 + VSE & 52.7 & 0.81 & 54.5 & 0.84\\
&CSN, T4:1 + VSE + Sim & 53.1 & 0.81 & 54.4 & 0.85\\
&CSN, T4:1 + VSE + Sim + Metric  & 53.7 & 0.82 & 55.1 & 0.85\\
\hline
{\bf (b)}&CSN, T1:1, FC & 53.3 & 0.82 & 54.6 & 0.85\\
&CSN, T1:1 + VSE, FC & 53.7 & 0.82 & 55.2 & 0.86\\
&CSN, T1:1 + VSE + Sim, FC & 53.7 & 0.83 & 55.6 & 0.86\\
&CSN, T1:1 + VSE + Sim + Metric, FC  & 54.0 & 0.83 & 56.6 & 0.86\\
\hline
{\bf (c)} & CSN, T1:1 + VSE + Sim + Metric & 54.1 & 0.82 & 55.3 & 0.86\\
&CSN, T1:1 + VSE + Sim + Cosine & 53.9 & 0.82 & 54.8 & 0.86\\
\hline
{\bf (d)} & CSN, T1:1 + Sim & 53.1 & 0.82 & 54.4 & 0.84\\
&CSN, T1:1 + Metric & 53.3 & 0.83 & 54.6 & 0.84\\
& CSN, T1:1 + Sim + Metric & 53.6 & 0.83 & 54.8 & 0.85\\
\hline
\end{tabular}
\end{table}

\section{Learned Compatibility Relationships}

The following figures show learned compatibility relationships with our model. The query item is show in the top row and is randomly pulled from some type $u$. Rows highlighted in yellow show our model's suggestions for compatible items to the query item of a randomly selected type $v$. Bottom rows show items of type $v$ sampled at random. 

\begin{figure}
\centering
\includegraphics[width=\textwidth]{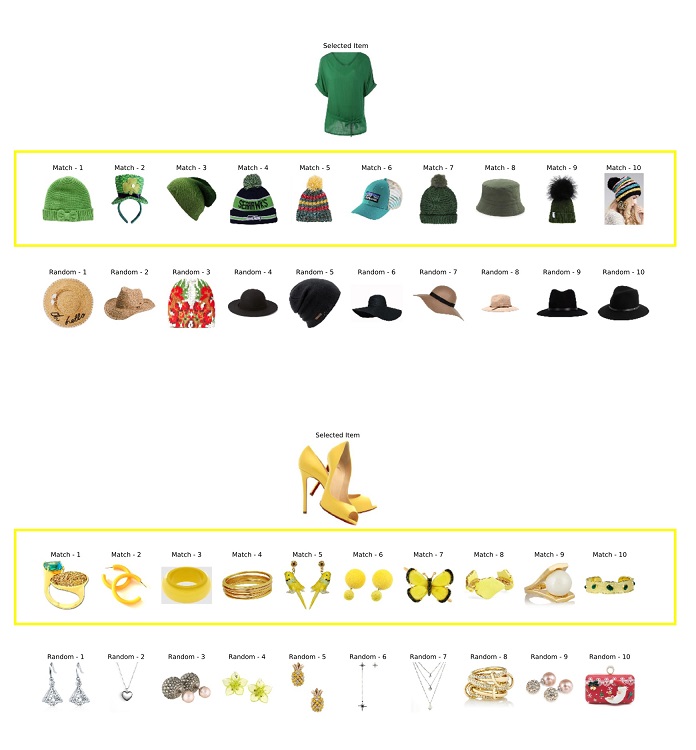}
\end{figure}

\begin{figure}
\centering
\includegraphics[width=\textwidth]{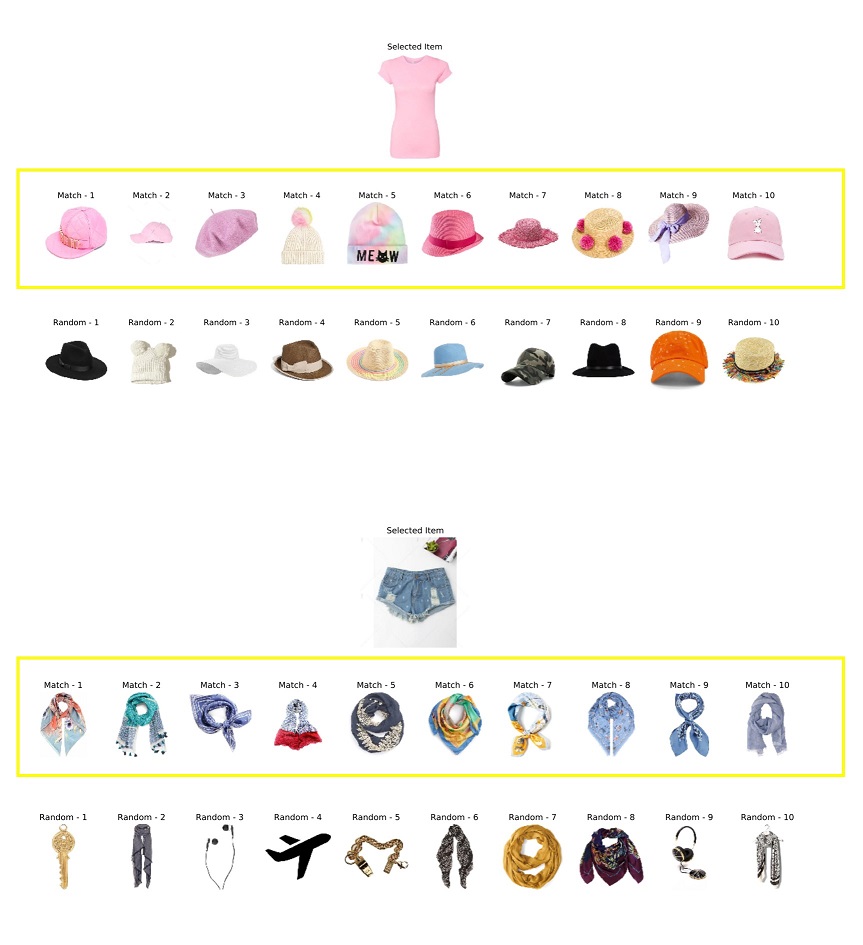}
\end{figure}

\begin{figure}
\centering
\includegraphics[width=\textwidth]{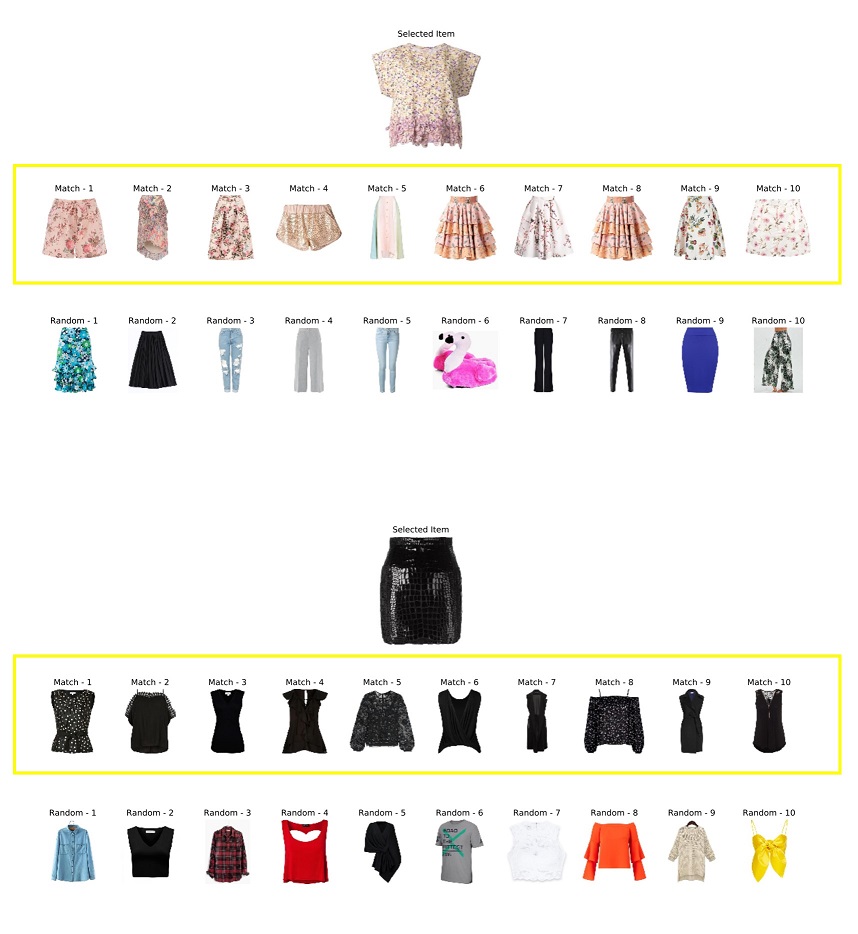}
\end{figure}

\begin{figure}
\centering
\includegraphics[width=\textwidth]{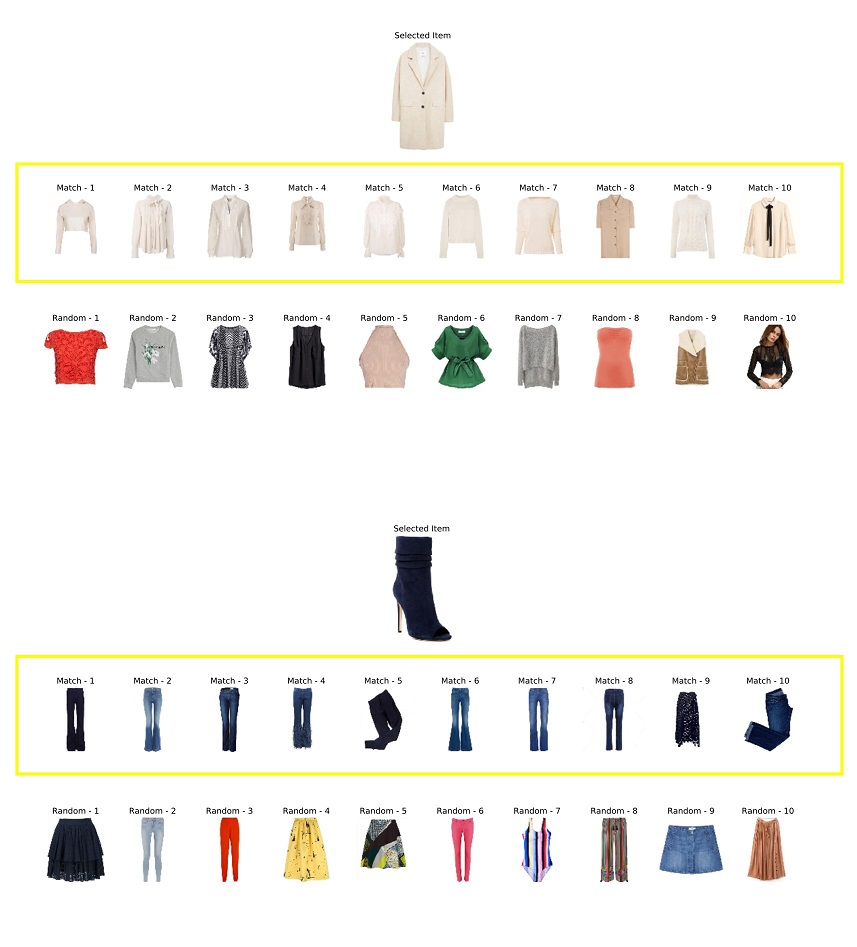}
\end{figure}

\begin{figure}
\centering
\includegraphics[width=\textwidth]{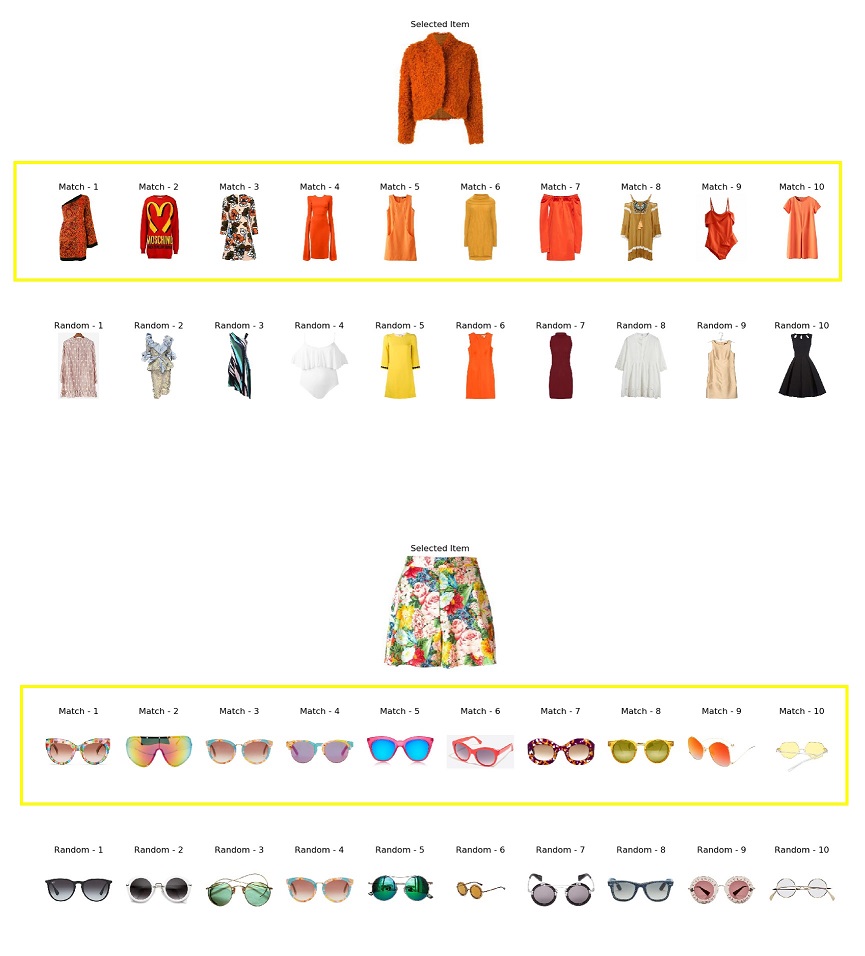}
\end{figure}

\begin{figure}
\centering
\includegraphics[width=\textwidth]{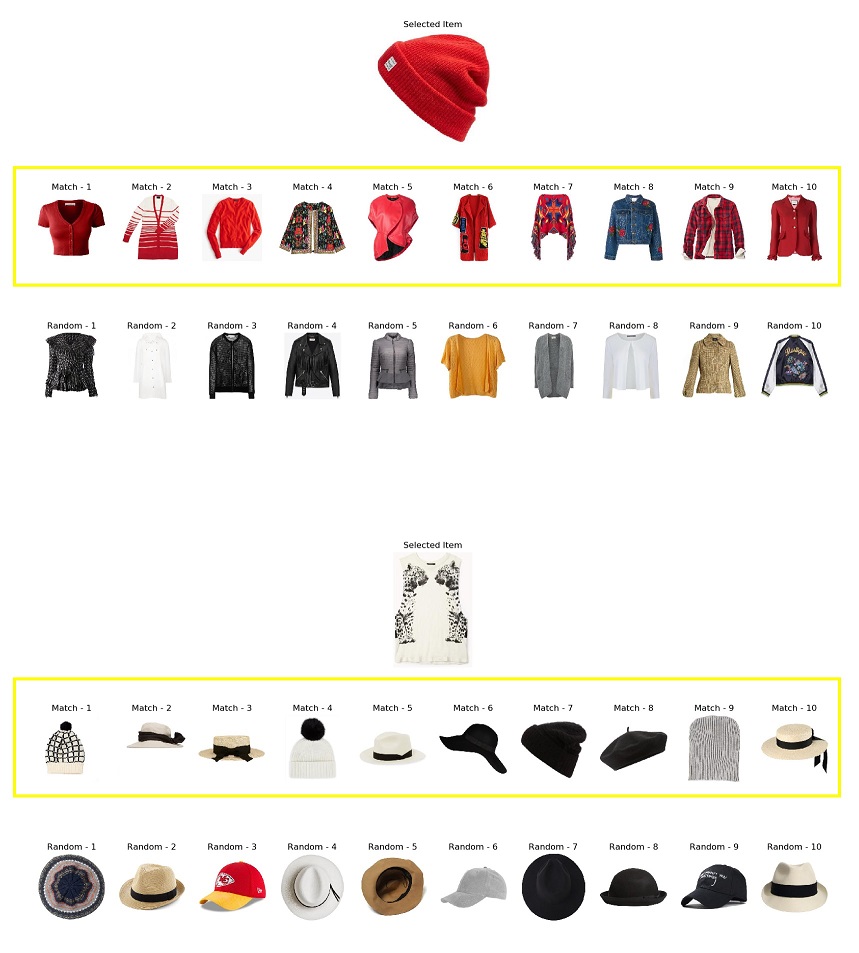}
\end{figure}

\begin{figure}
\centering
\includegraphics[width=\textwidth]{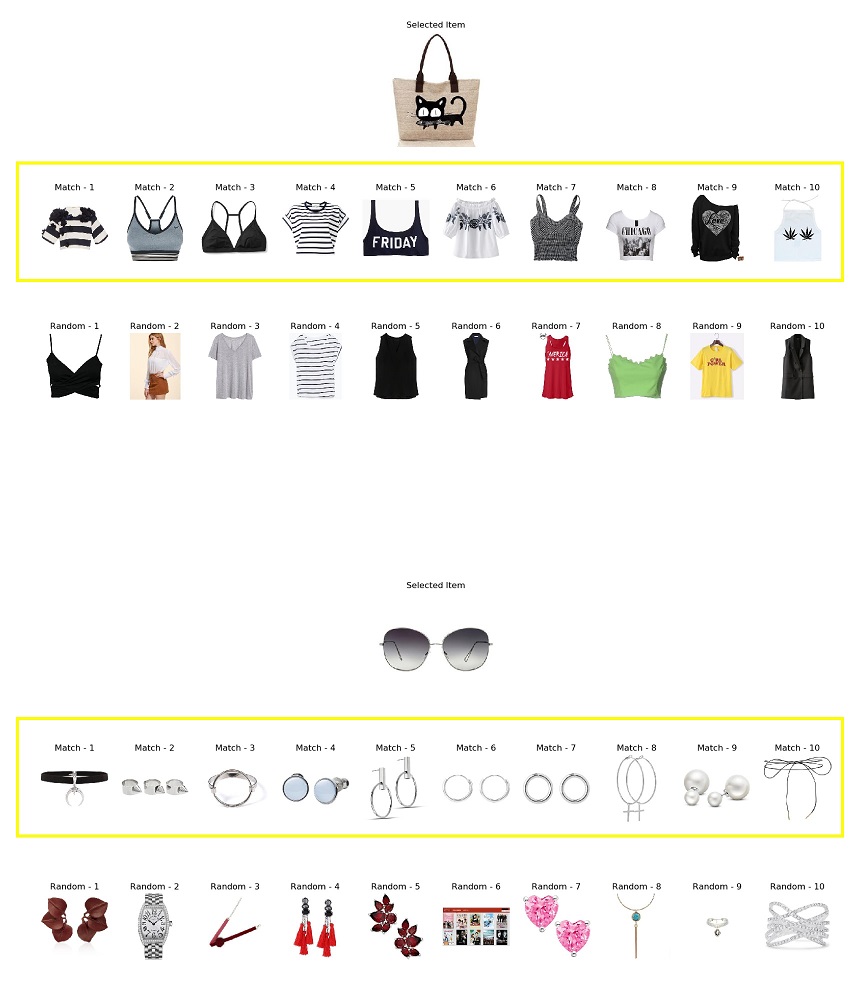}
\end{figure}

\begin{figure}
\centering
\includegraphics[width=\textwidth]{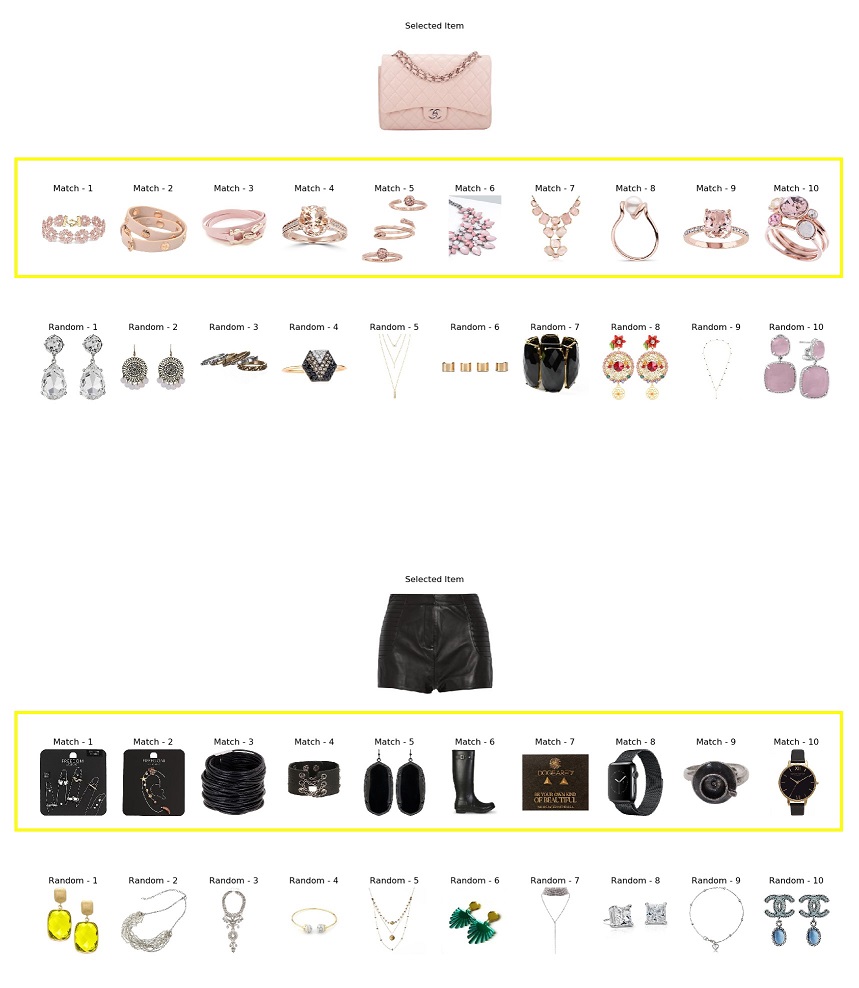}
\end{figure}

\begin{figure}
\centering
\includegraphics[width=\textwidth]{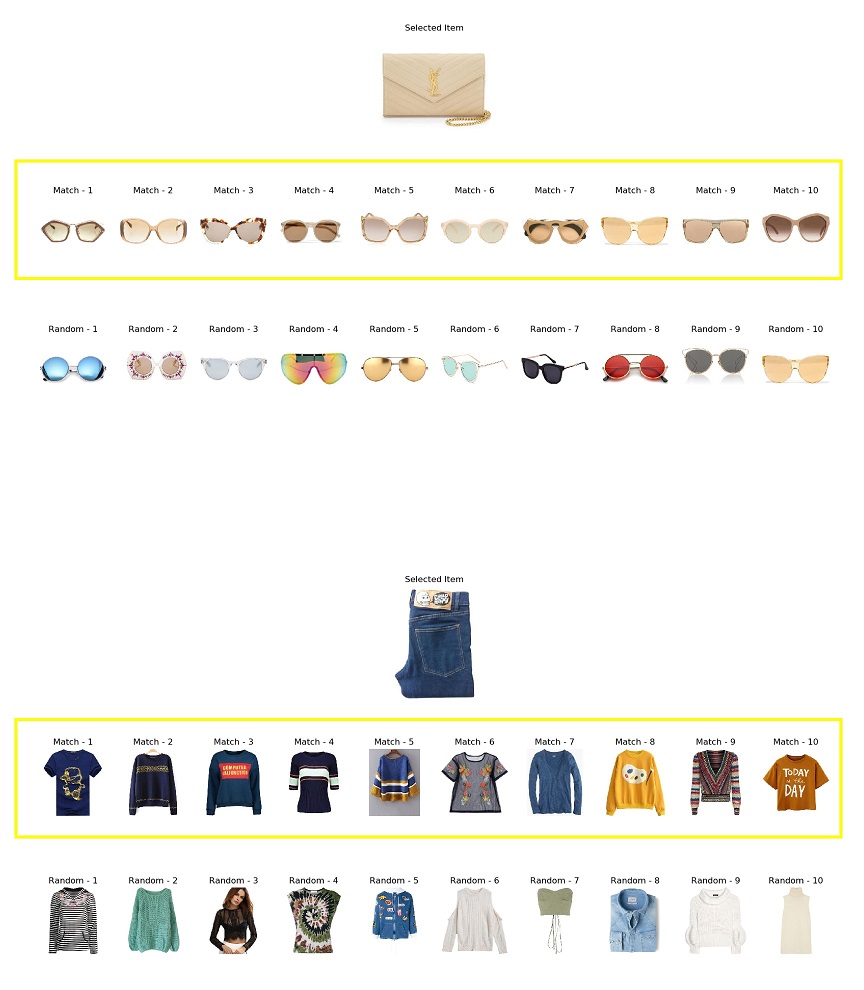}
\end{figure}

\begin{figure}
\centering
\includegraphics[width=\textwidth]{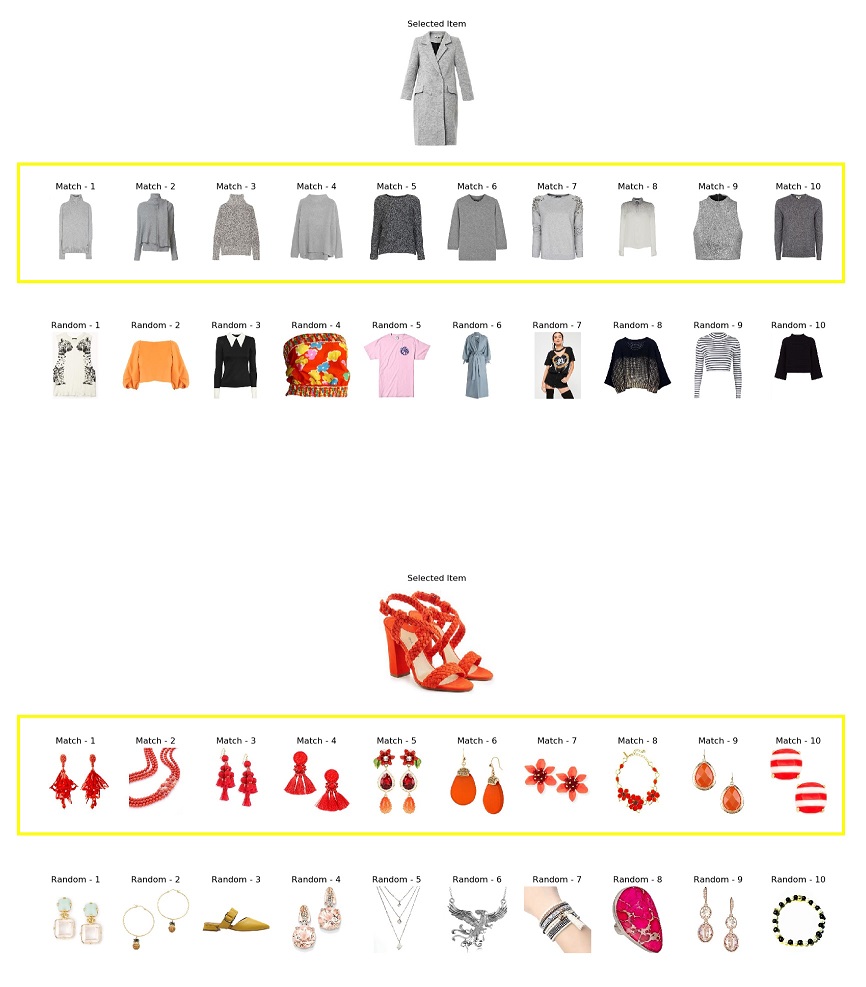}
\end{figure}

\begin{figure}
\centering
\includegraphics[width=\textwidth]{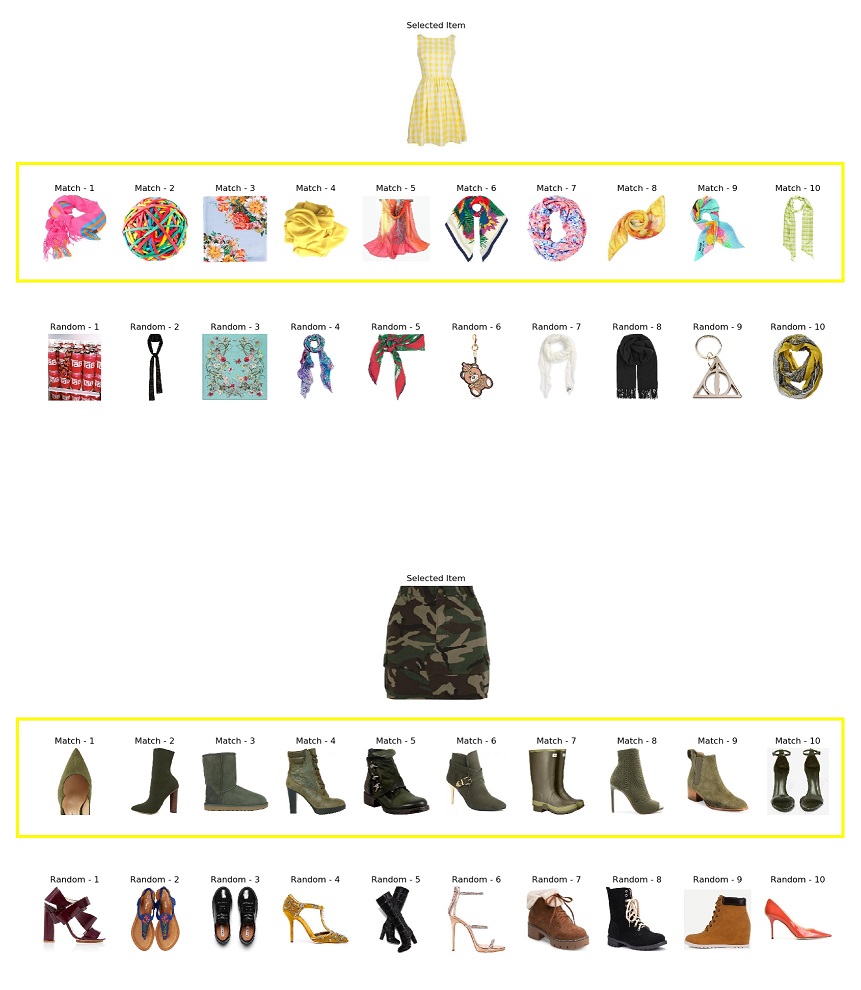}
\end{figure}

\section{Learned Similarity Relationships}

The following figures show examples of learned similarity relationships (\ie items which are generally considered alternatives in an outfit) by our model. Columns highlighted in yellow show the query item for each row. Rows contain similar items of the same type as the query item. 

\begin{figure}
\centering
\includegraphics[width=\textwidth]{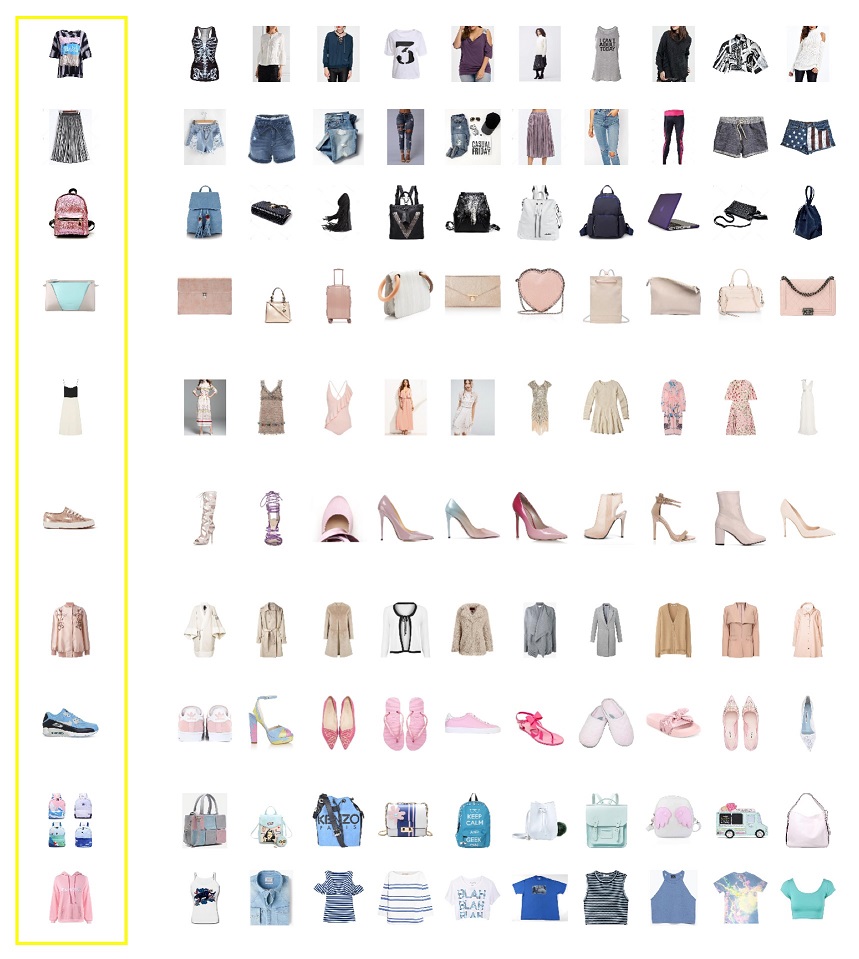}
\end{figure}

\begin{figure}
\centering
\includegraphics[width=\textwidth]{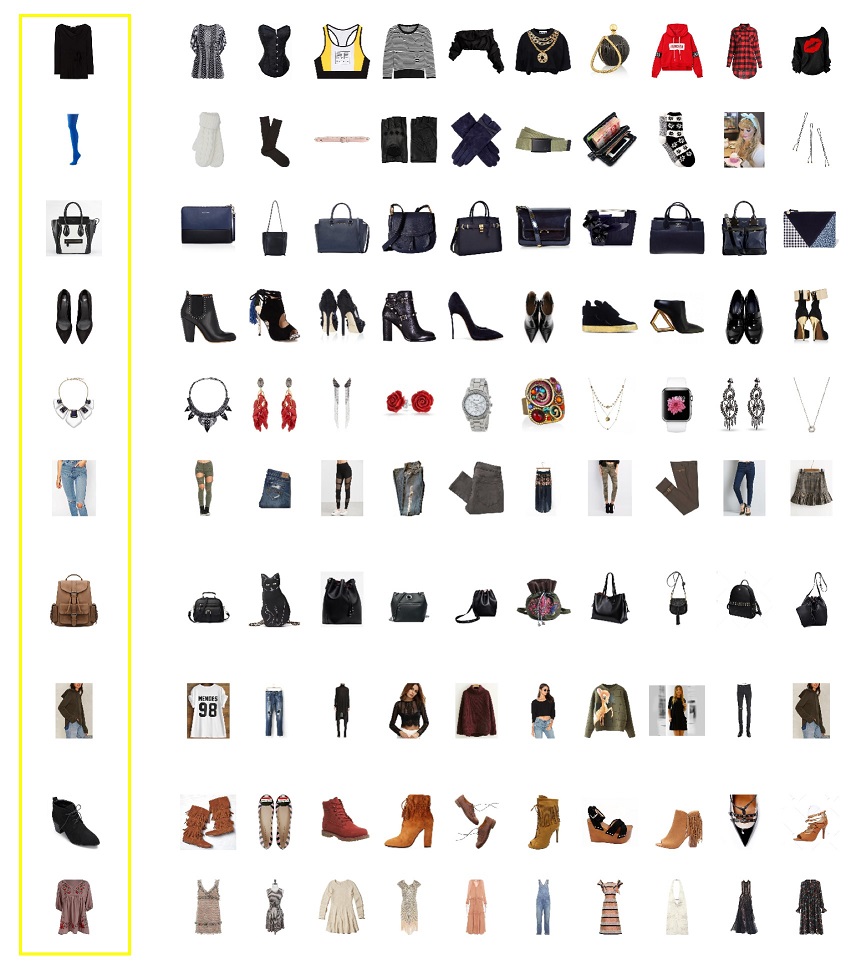}
\end{figure}

\begin{figure}
\centering
\includegraphics[width=\textwidth]{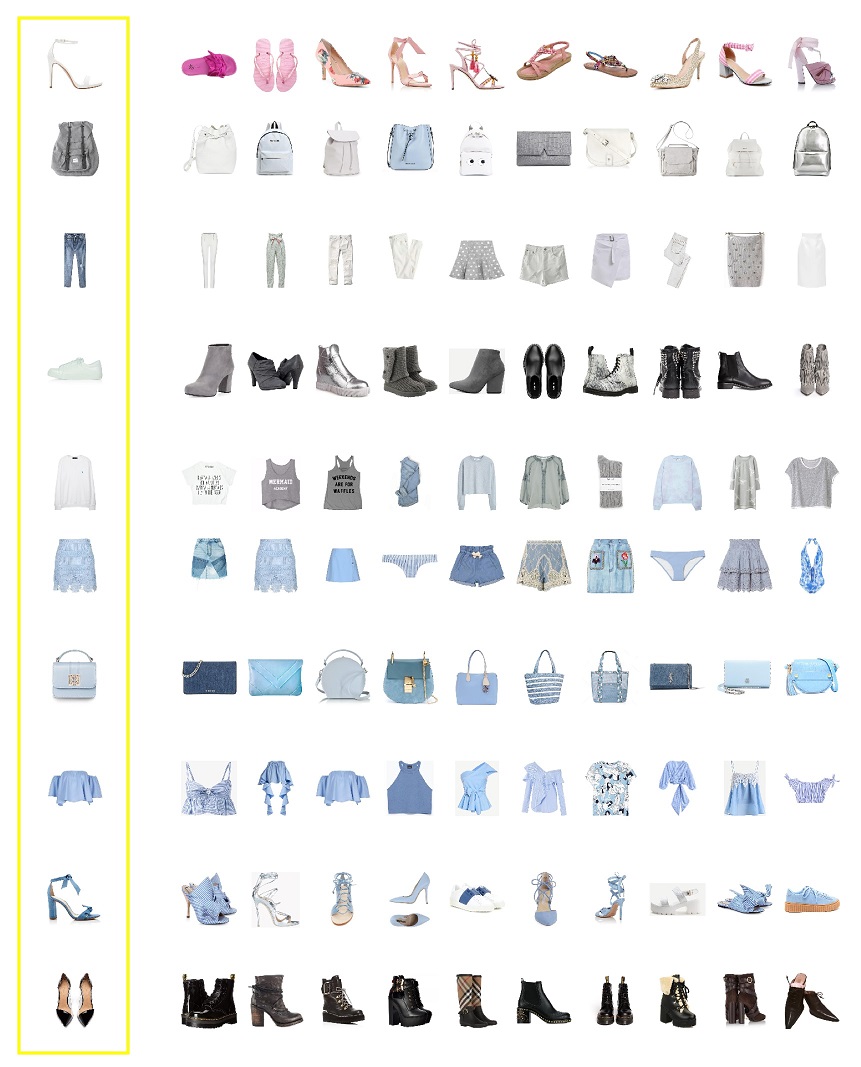}
\end{figure}

\begin{figure}
\centering
\includegraphics[width=\textwidth]{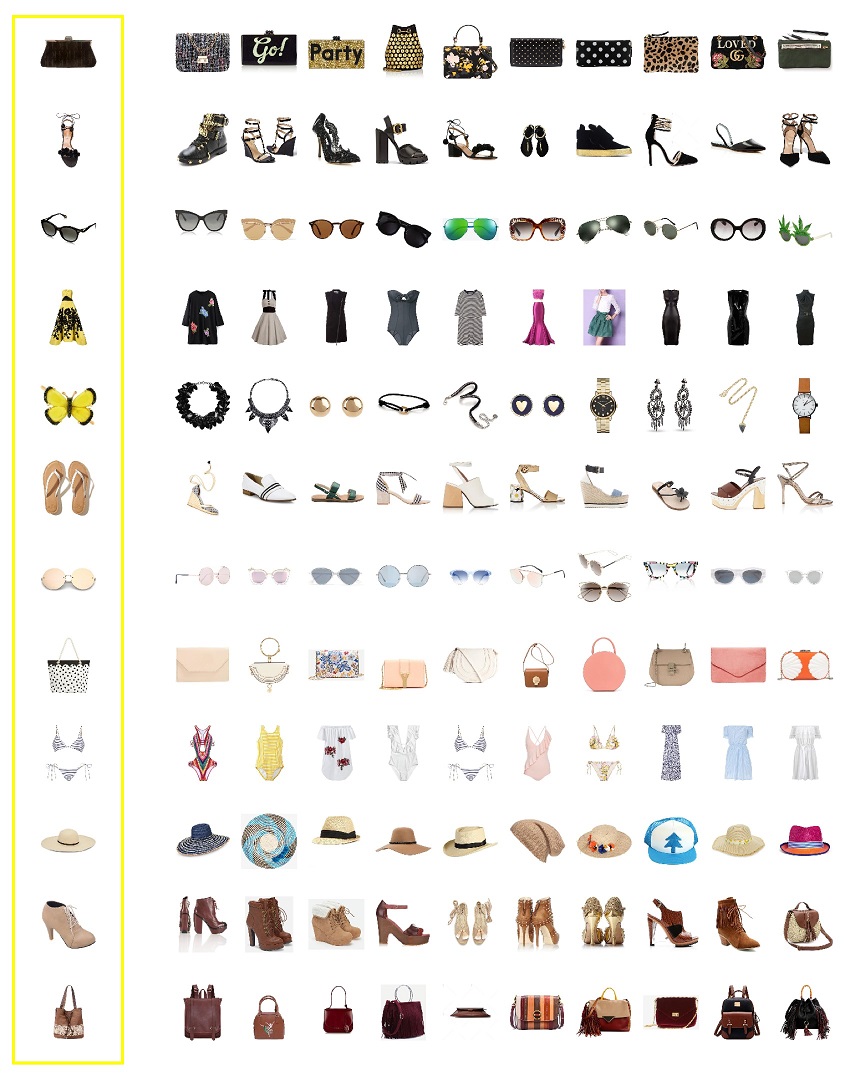}
\end{figure}

\begin{figure}
\centering
\includegraphics[width=\textwidth]{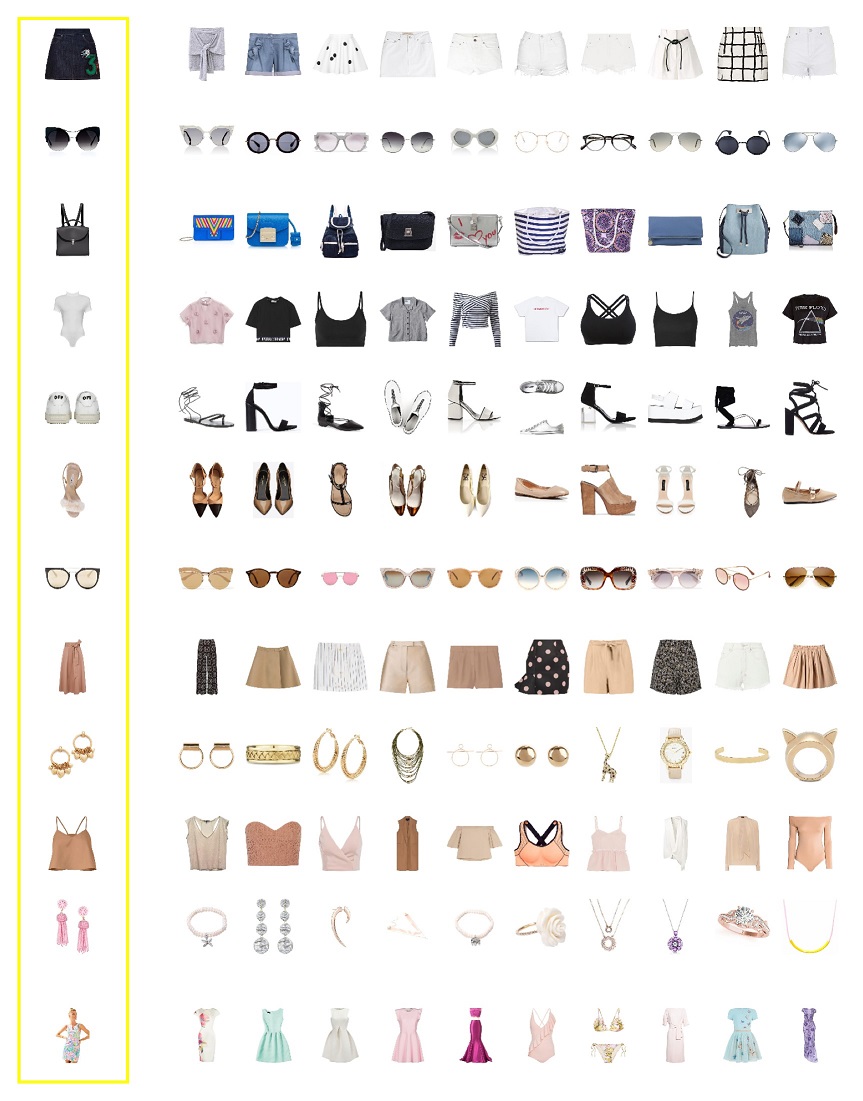}
\end{figure}

\begin{figure}
\centering
\includegraphics[width=\textwidth]{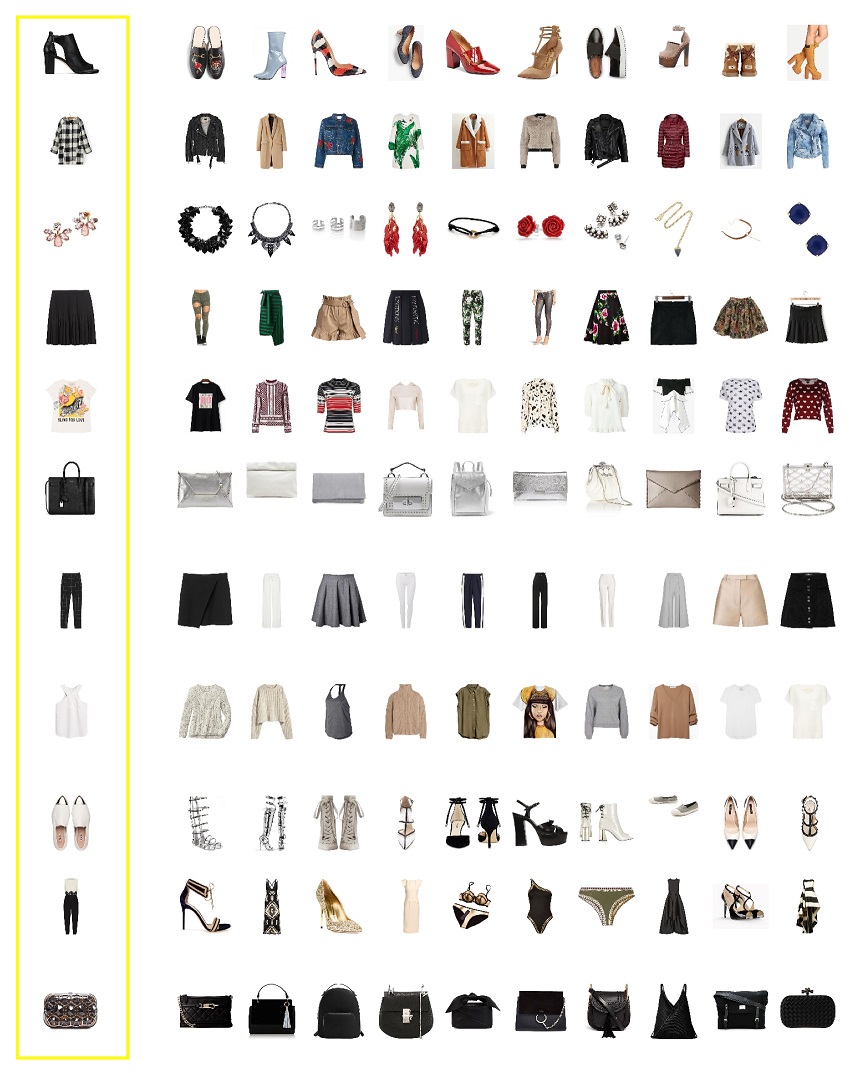}
\end{figure}

\begin{figure}
\centering
\includegraphics[width=\textwidth]{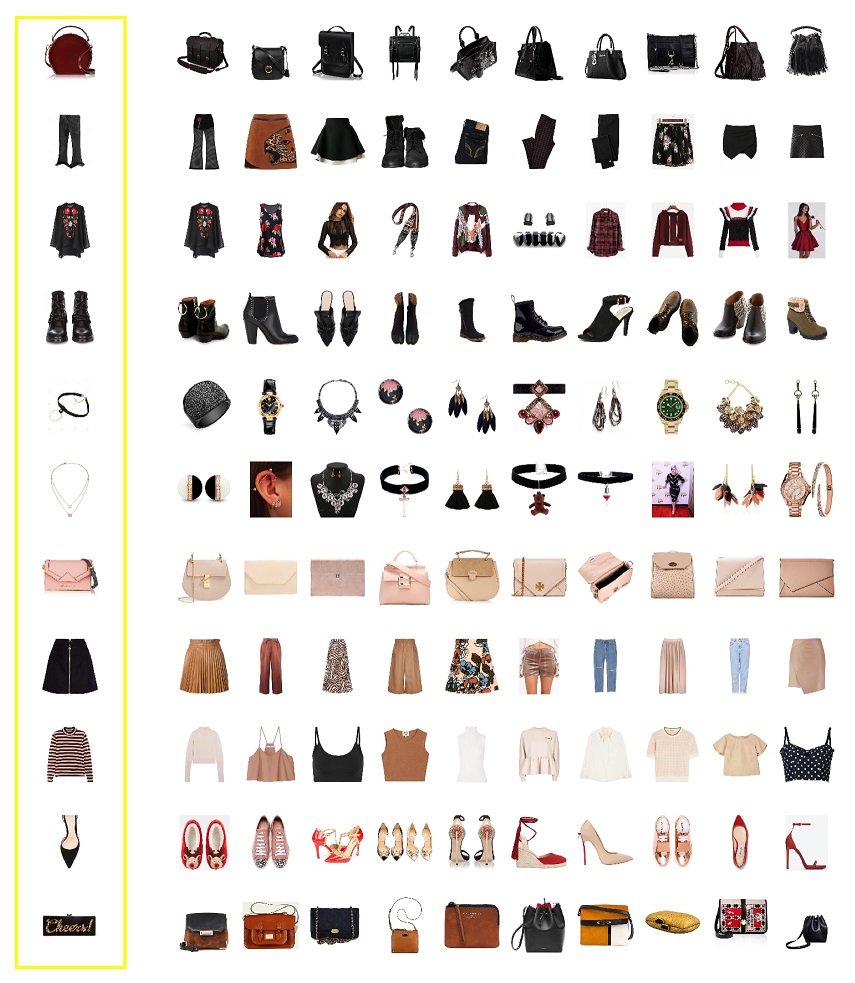}
\end{figure}

\begin{figure}
\centering
\includegraphics[width=\textwidth]{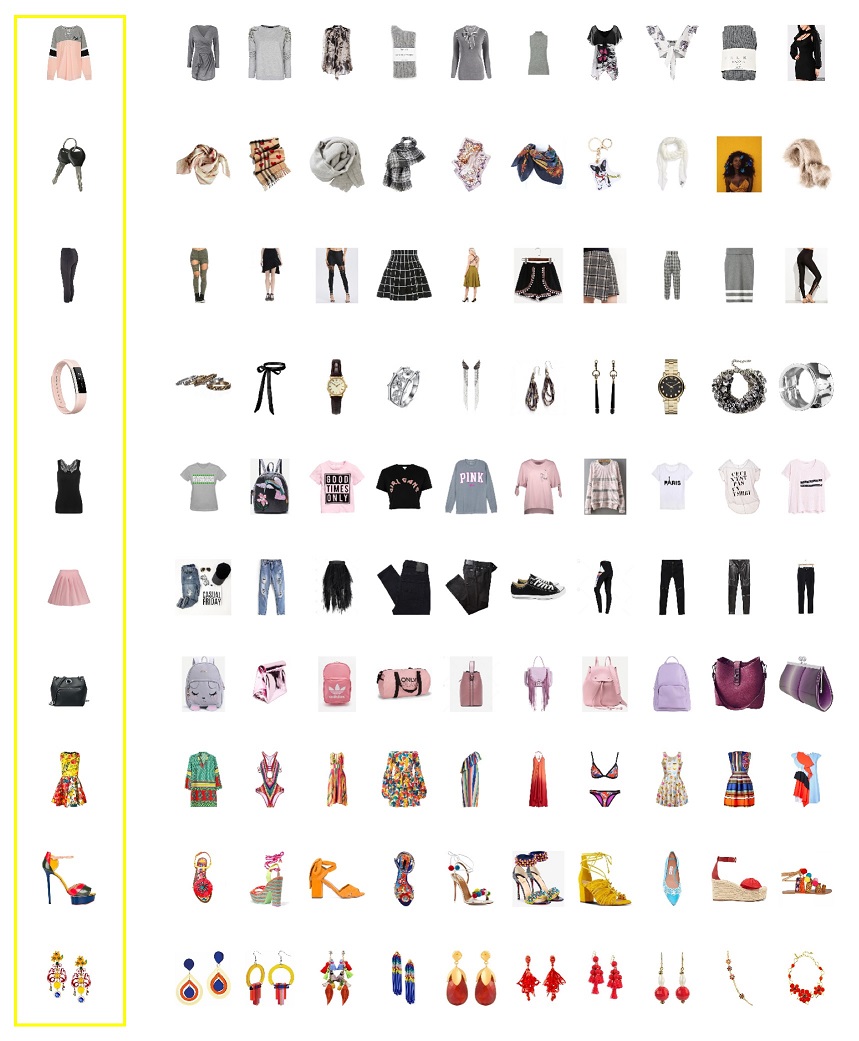}
\end{figure}

\begin{figure}
\centering
\includegraphics[width=\textwidth]{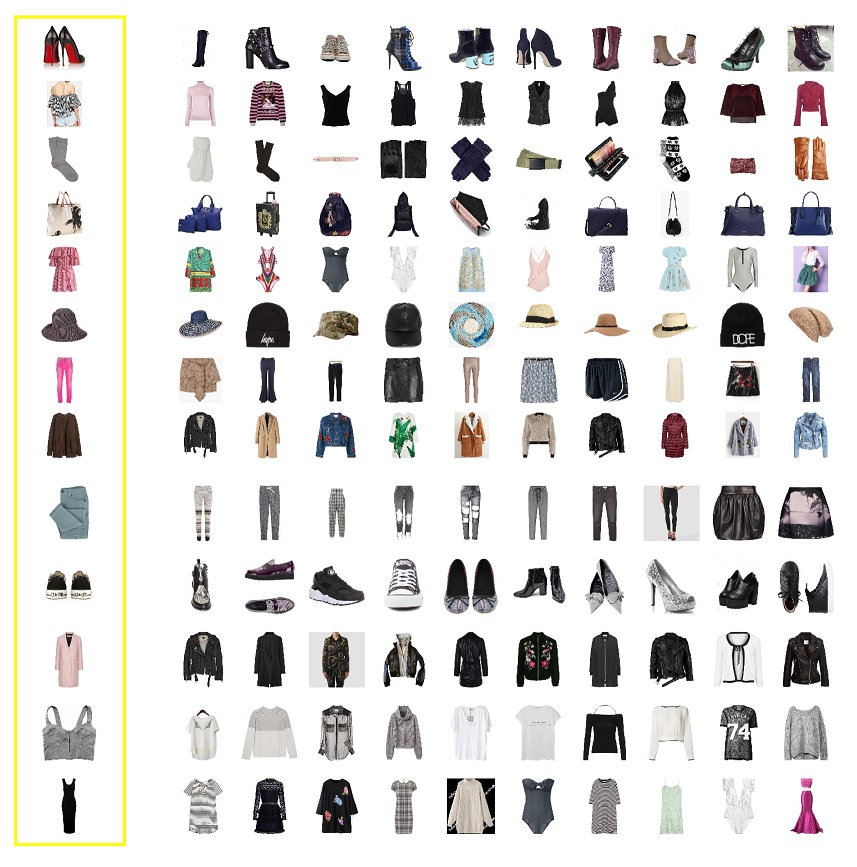}
\end{figure}

\begin{figure}
\centering
\includegraphics[width=\textwidth]{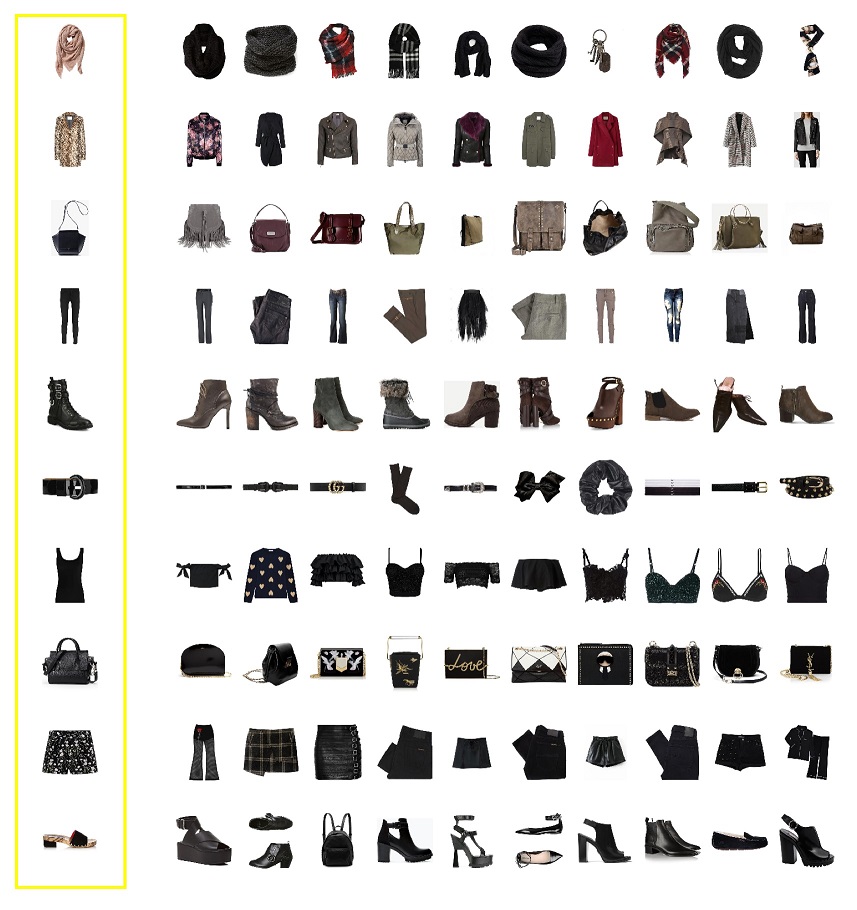}
\end{figure}

\begin{figure}
\centering
\includegraphics[width=\textwidth]{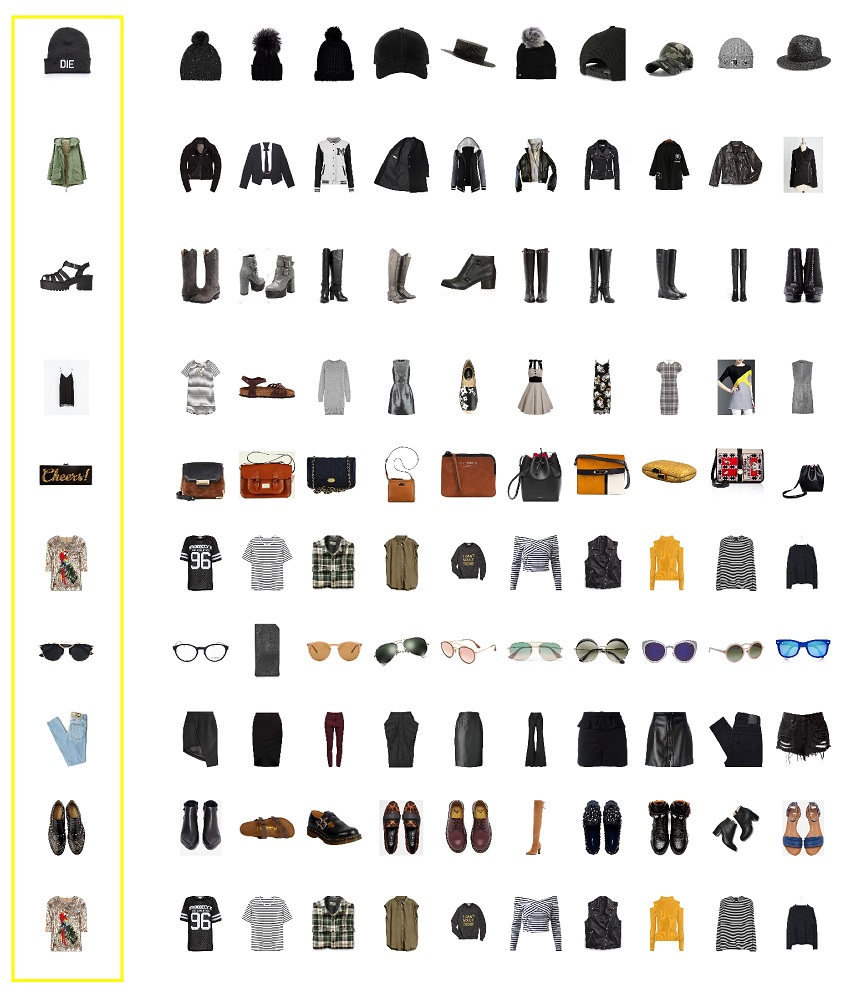}
\end{figure}

\section{Outfit Diversification by Item Replacement}

The following figures show examples of outfit diversification by replacement of a single item. Top rows represent a valid (i.e., human-curated) outfit. Highlighted in blue is a randomly selected heldout item from the valid outfit to be replaced. Highlighted in green are items of the valid outfit that remain unchanged. Rows highlighted in yellow display alternatives to the heldout item that are all equally compatible with the rest of the outfit (i.e., with the collection of items in green) according to our model. Bottorm rows show a random selection of alternatives of the same type as the heldout item. 

\begin{figure}
\centering
\includegraphics[width=0.92\textwidth]{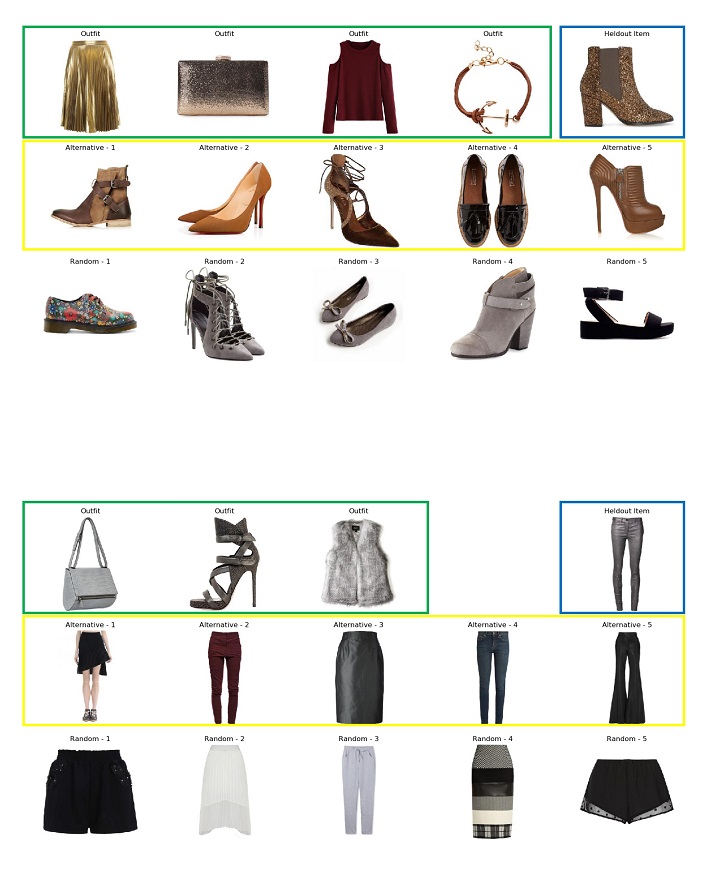}
\end{figure}

\begin{figure}
\centering
\includegraphics[width=\textwidth]{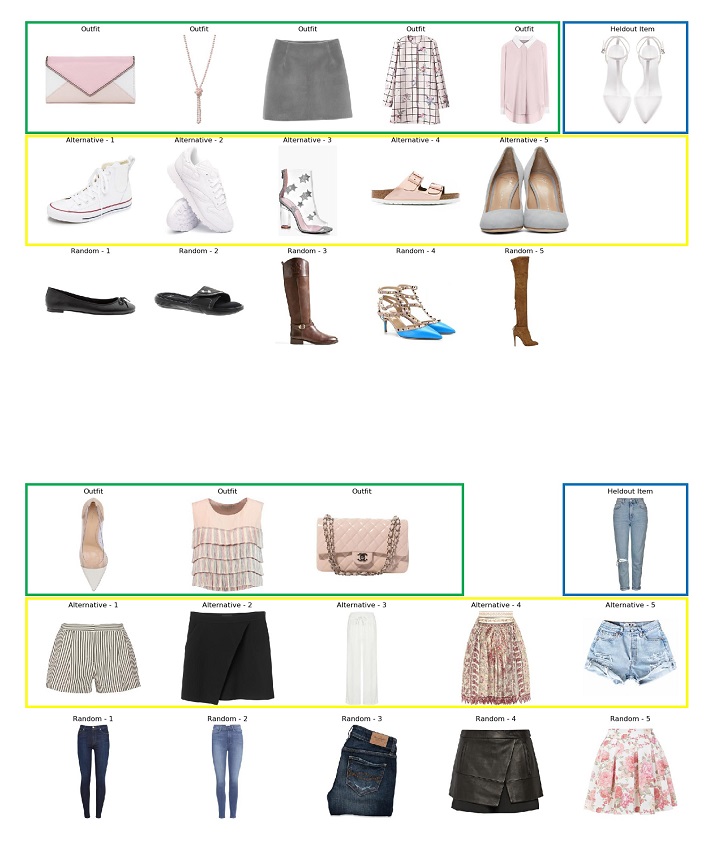}
\end{figure}

\begin{figure}
\centering
\includegraphics[width=\textwidth]{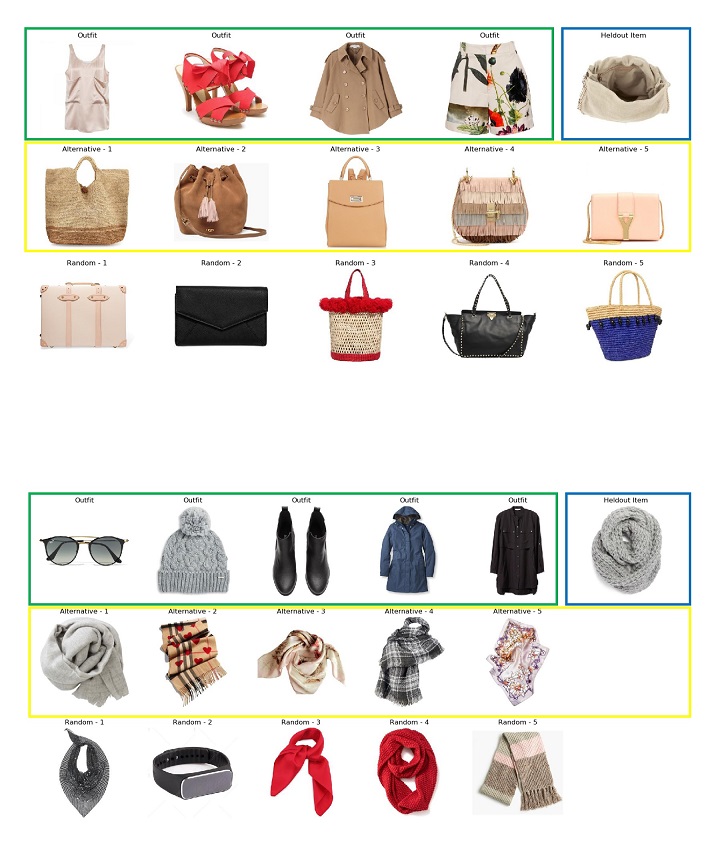}
\end{figure}

\begin{figure}
\centering
\includegraphics[width=\textwidth]{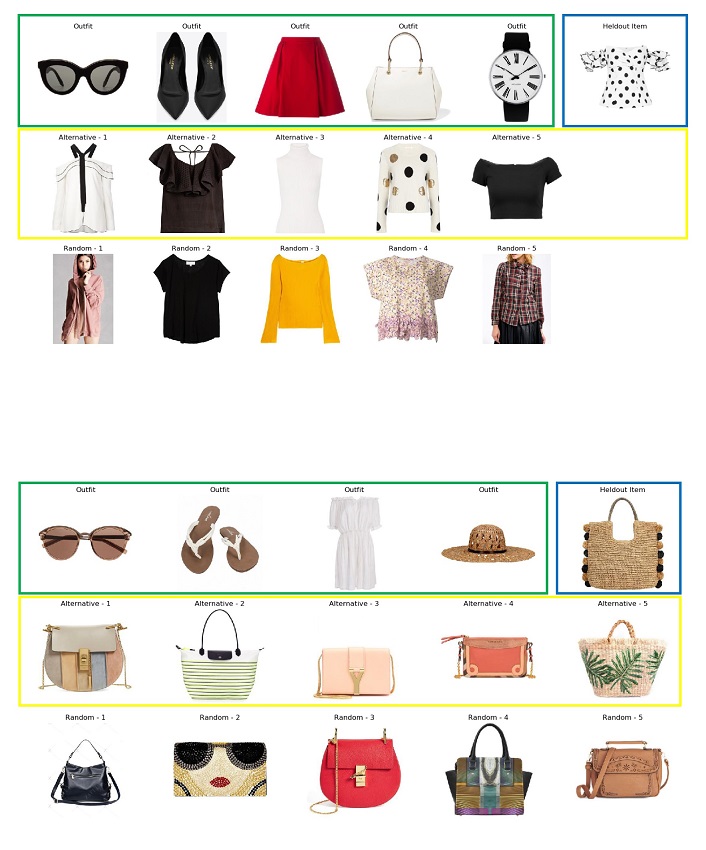}
\end{figure}

\begin{figure}
\centering
\includegraphics[width=\textwidth]{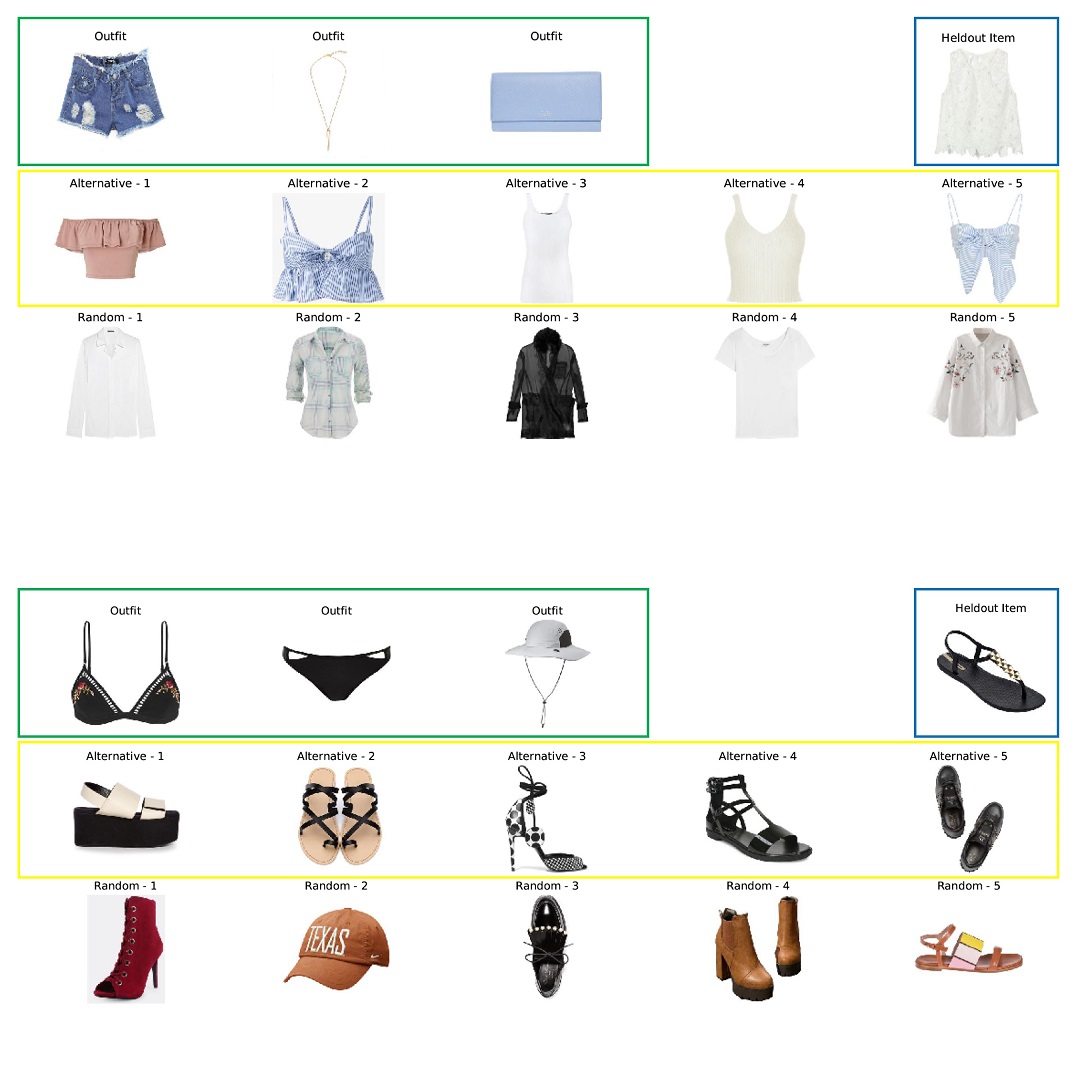}
\end{figure}

\begin{figure}
\centering
\includegraphics[width=\textwidth]{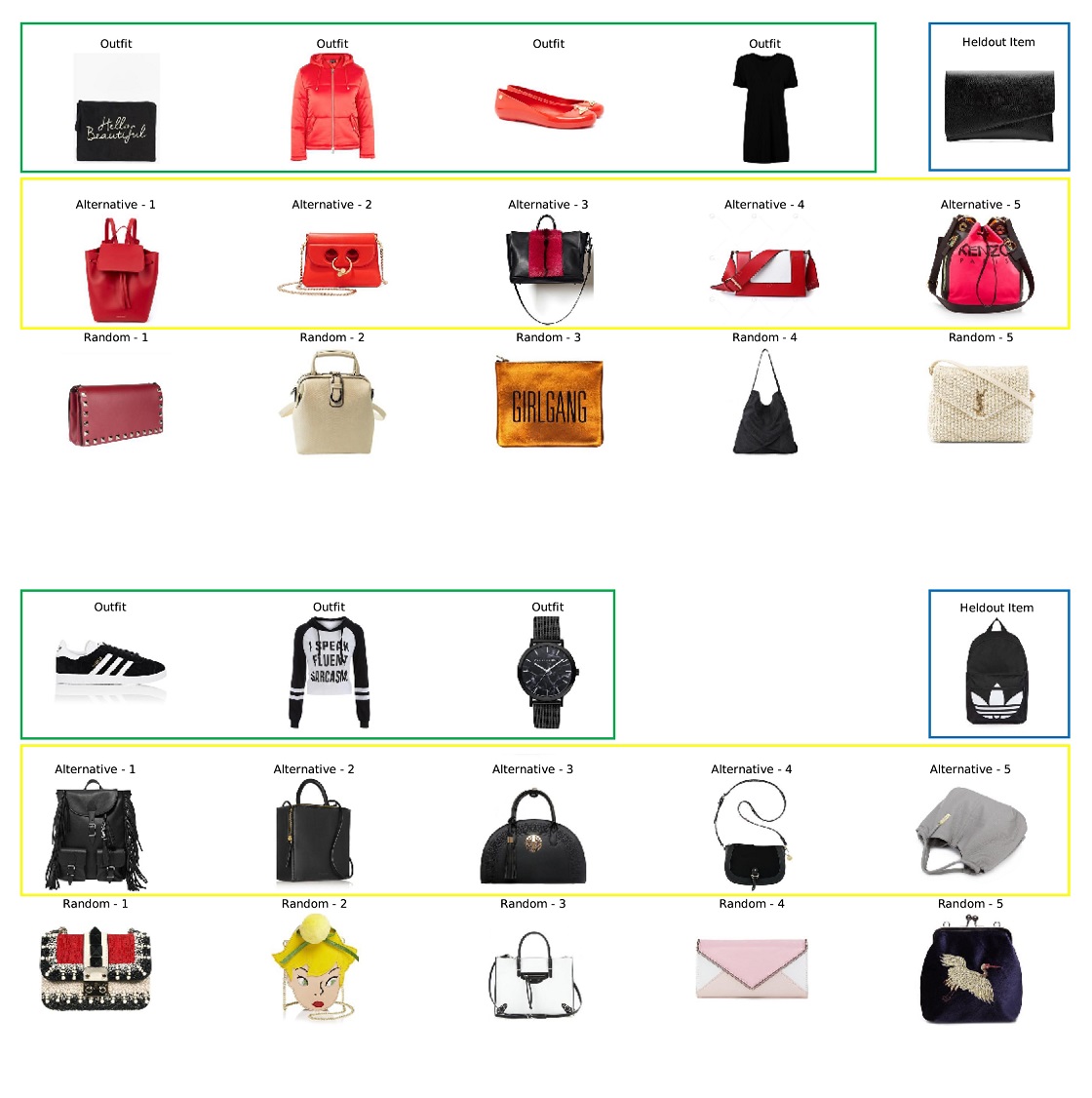}
\end{figure}

\begin{figure}
\centering
\includegraphics[width=\textwidth]{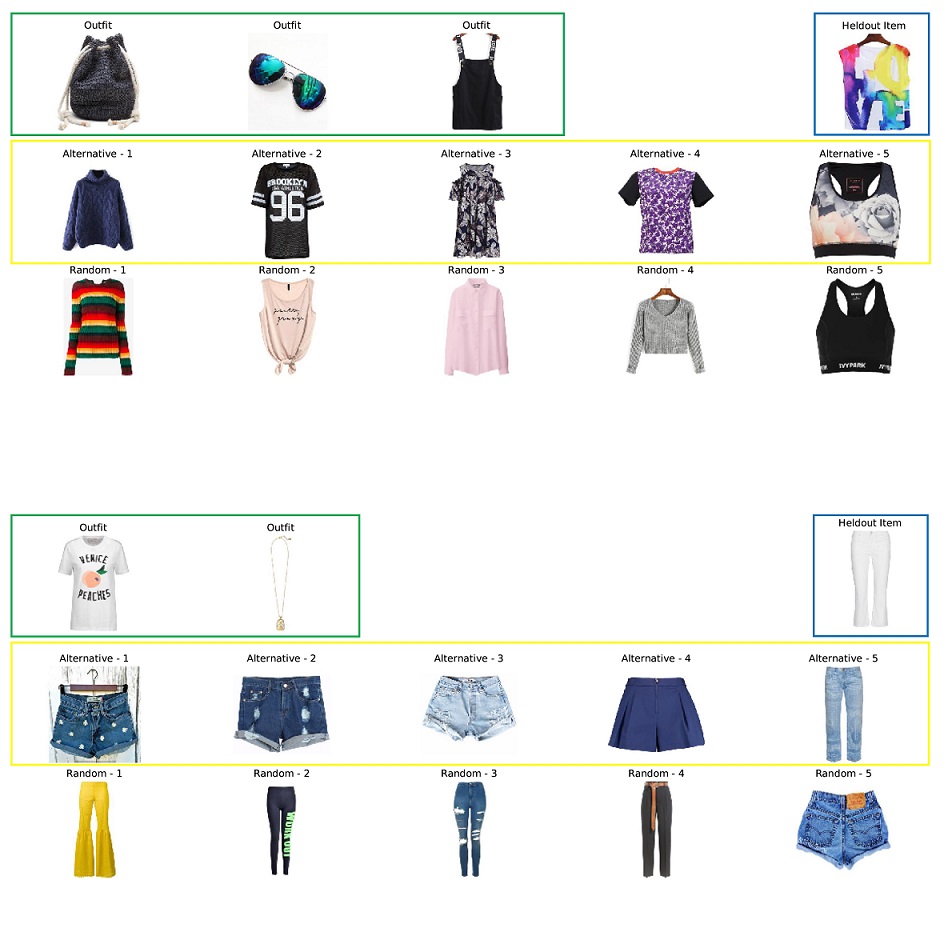}
\end{figure}

\section{Outfit Generation by Recursive Item Swaps}

The following figures show examples of outfit generation by recursive item swaps. The top row represents a valid (i.e., human-curated) outfit. At each step, we replace an item from the starting outfit with one that is of the same type and equally compatible with the rest of the outfit, but different from the removed item. Each row corresponds to one step, and the boxed item is to be replaced.

\begin{figure}
\centering
\includegraphics[width=\textwidth]{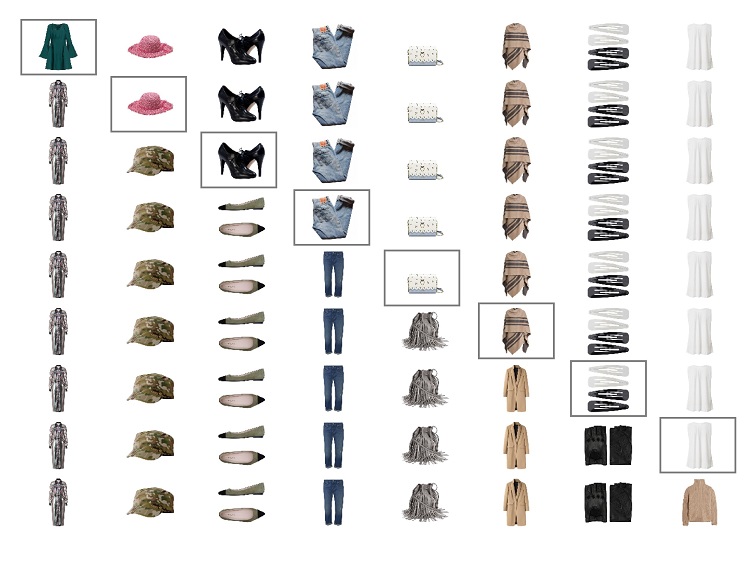}
\end{figure}

\begin{figure}
\centering
\includegraphics[height=\textheight]{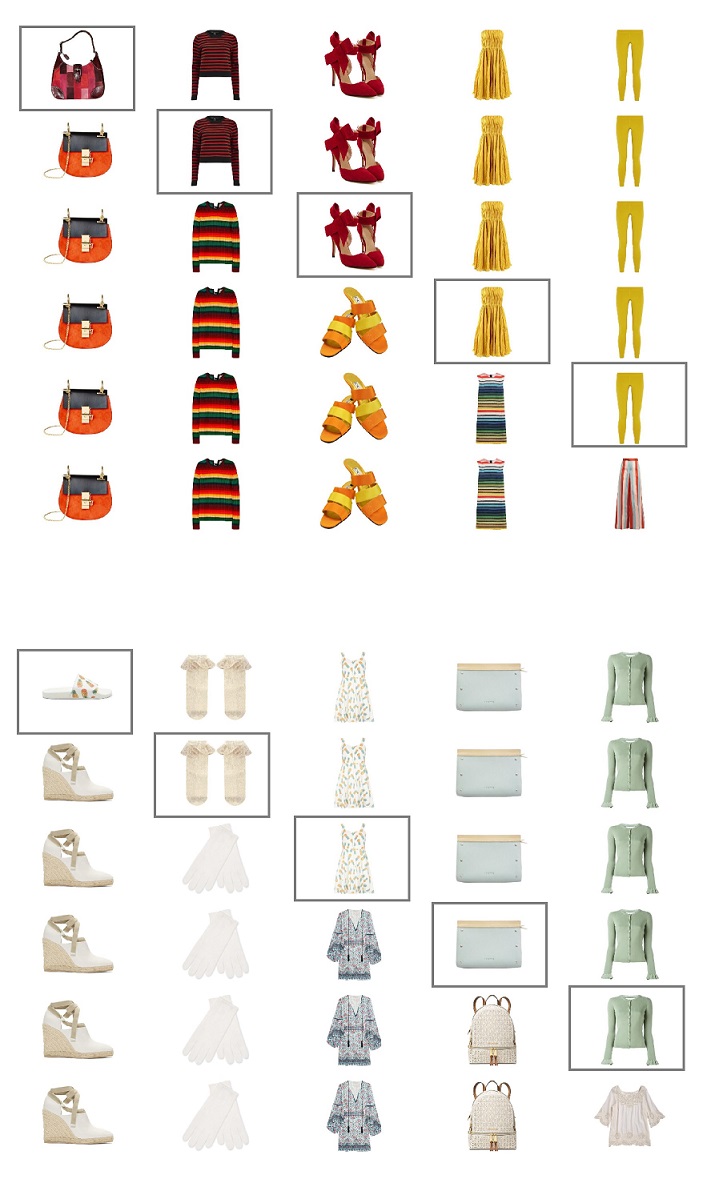}
\end{figure}

\begin{figure}
\centering
\includegraphics[height=\textheight]{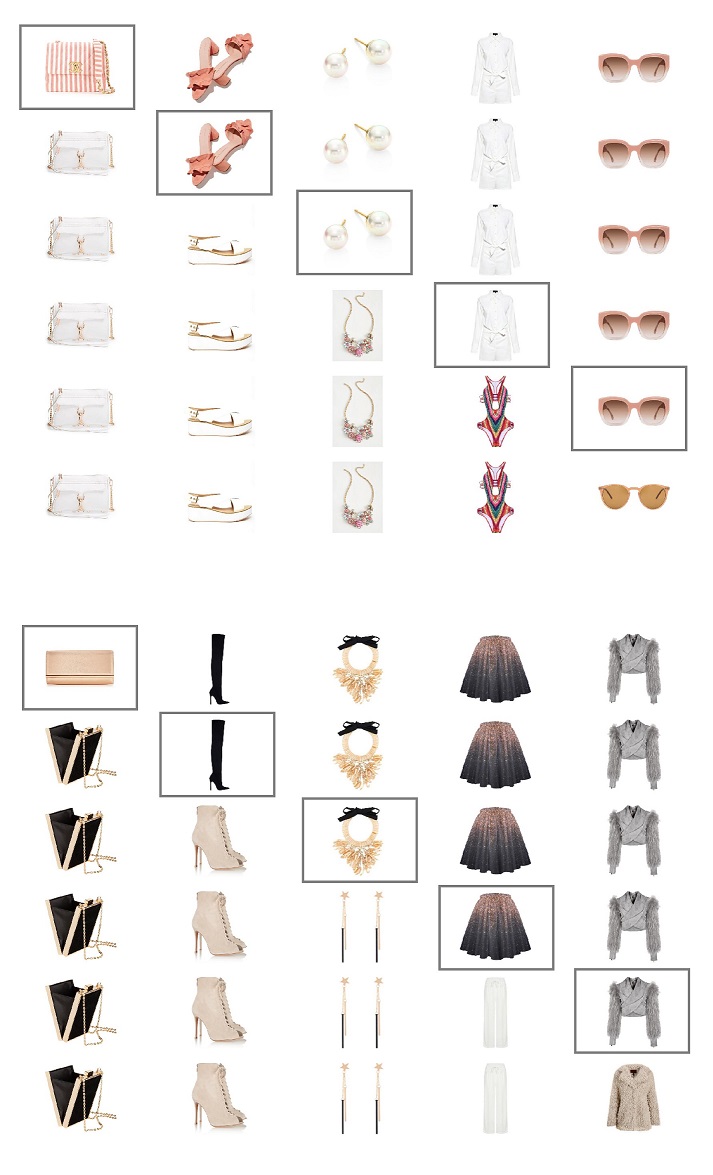}
\end{figure}

\begin{figure}
\centering
\includegraphics[height=\textheight]{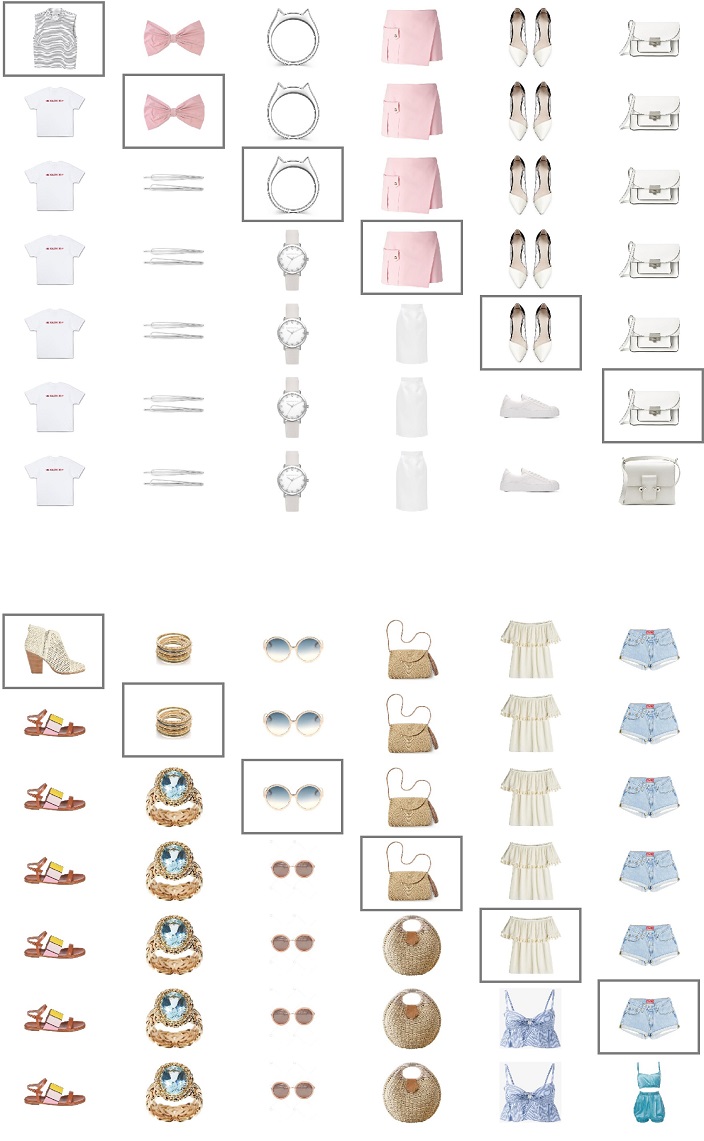}
\end{figure}

\begin{figure}
\centering
\includegraphics[width=\textwidth]{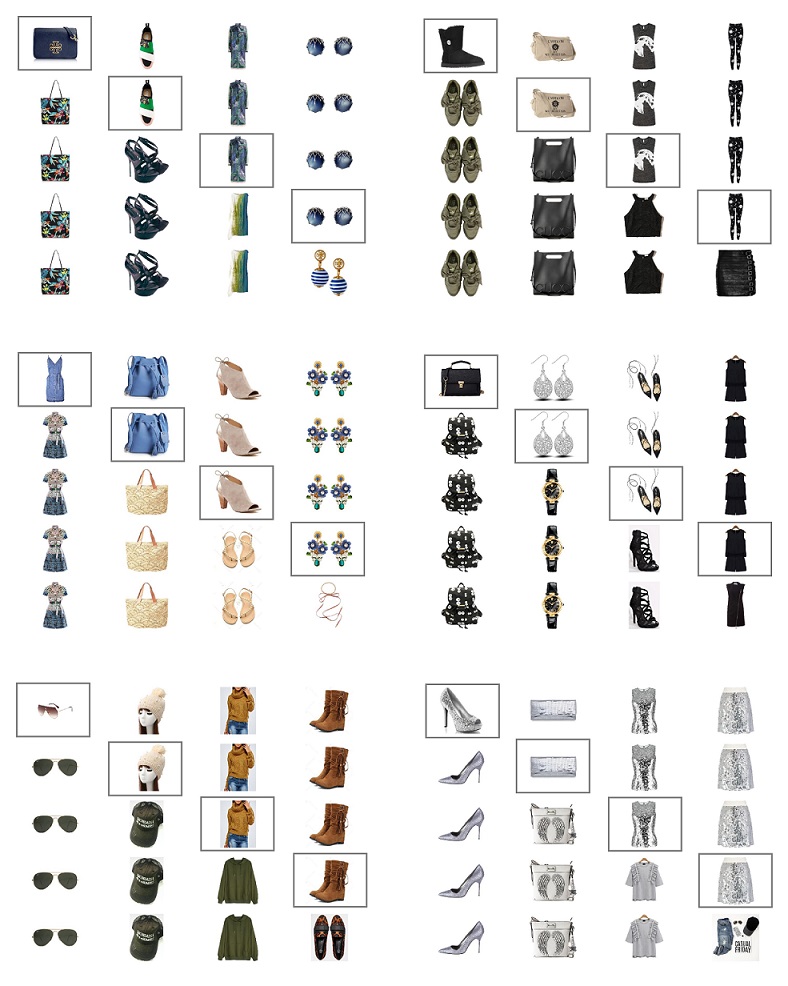}
\end{figure}

\begin{figure}
\centering
\includegraphics[height=\textheight]{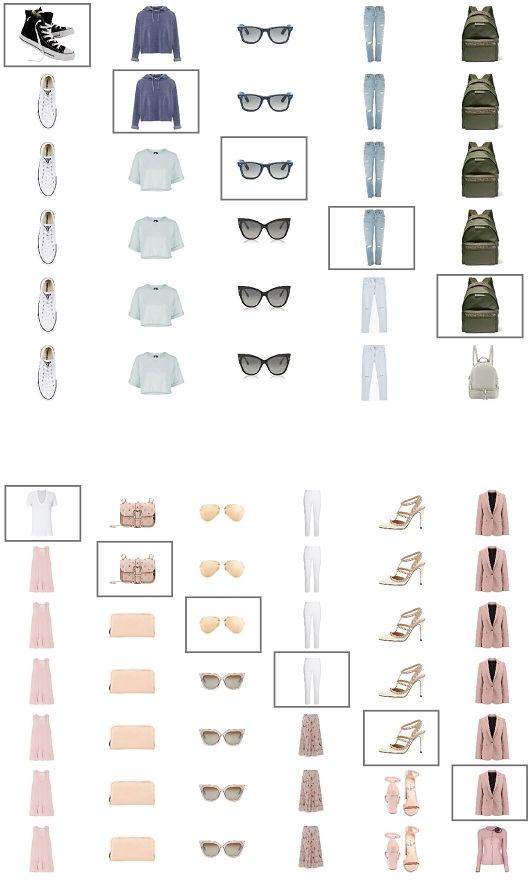}
\end{figure}

\begin{figure}
\centering
\includegraphics[width=\textwidth]{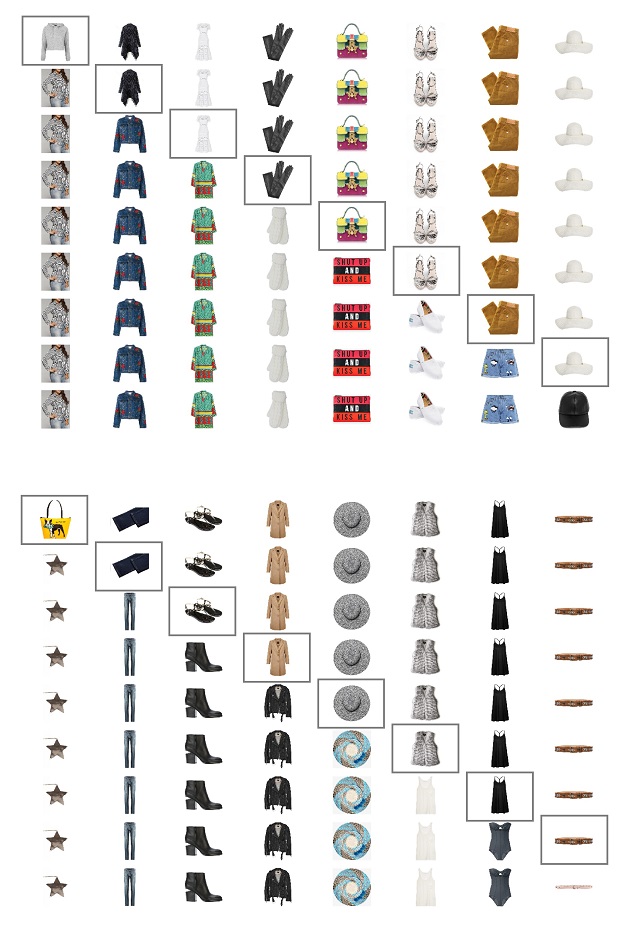}
\end{figure}

\begin{figure}
\centering
\includegraphics[height=0.9\textheight]{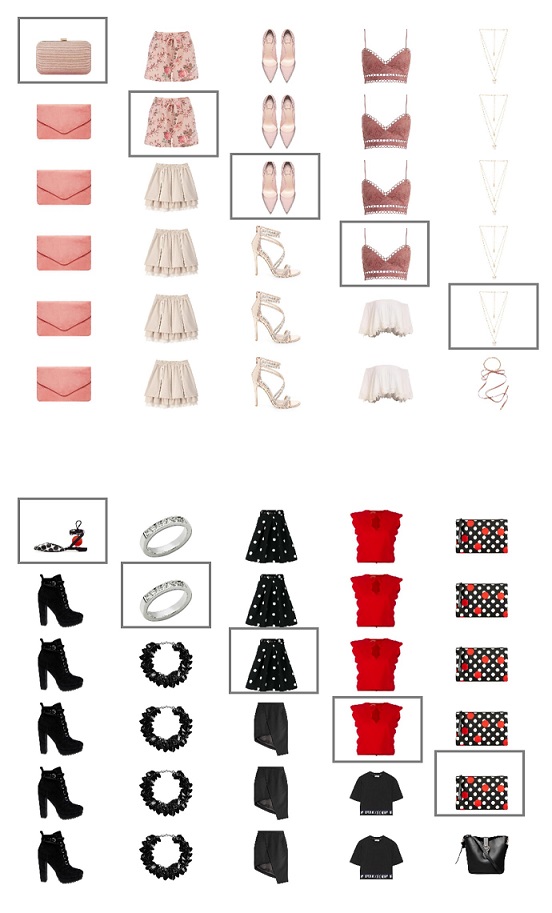}
\caption{\em \small Full figure showing outfit diversification by recursive item swaps used as examples in main paper}
\end{figure}

\section{Interpretability of Learned General Embedding}

\begin{figure}
\centering
\includegraphics[width=\textwidth]{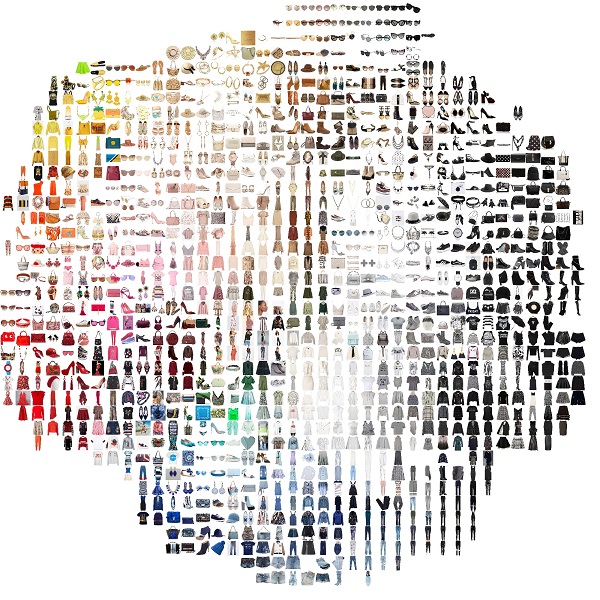}
\caption{\em \small t-SNE plot of the learned general embedding space on our Polyvore dataset}
\end{figure}

\clearpage

\section{Interpretability of Learned Type-Specific Embedding}
\begin{figure}
\centering
\includegraphics[width=\textwidth]{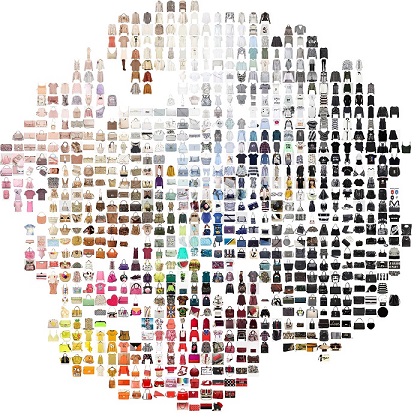}
\caption{\em \small t-SNE plot of the learned type-specific embedding space on our Polyvore dataset for tops and bags}
\end{figure}

\begin{figure}
\centering
\includegraphics[width=\textwidth]{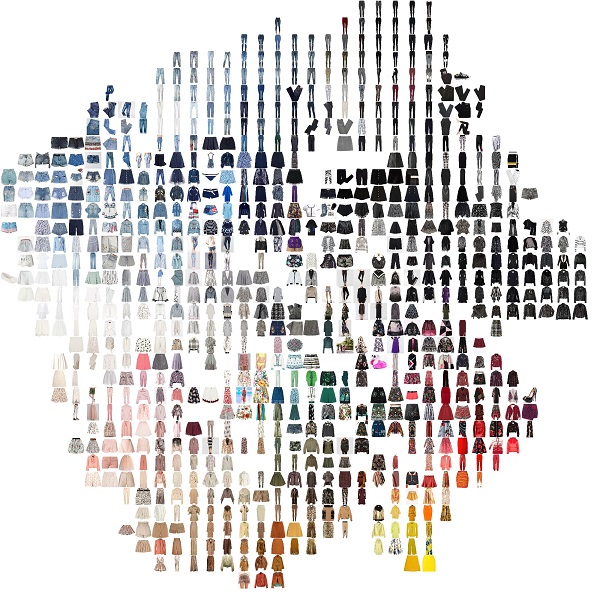}
\caption{\em \small t-SNE plot of the learned type-specific embedding space on our Polyvore dataset for bottoms and outerwear}
\end{figure}

\begin{figure}
\centering
\includegraphics[width=\textwidth]{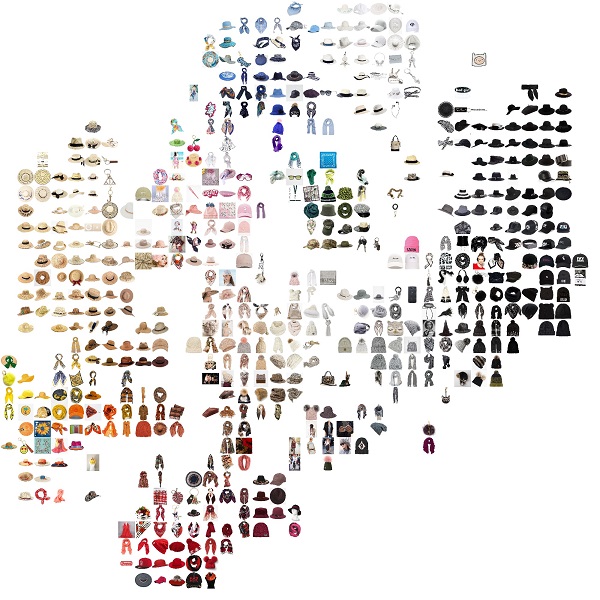}
\caption{\em \small t-SNE plot of the learned type-specific embedding space on our Polyvore dataset for hats and scarves}
\end{figure}

\begin{figure}
\centering
\includegraphics[width=\textwidth]{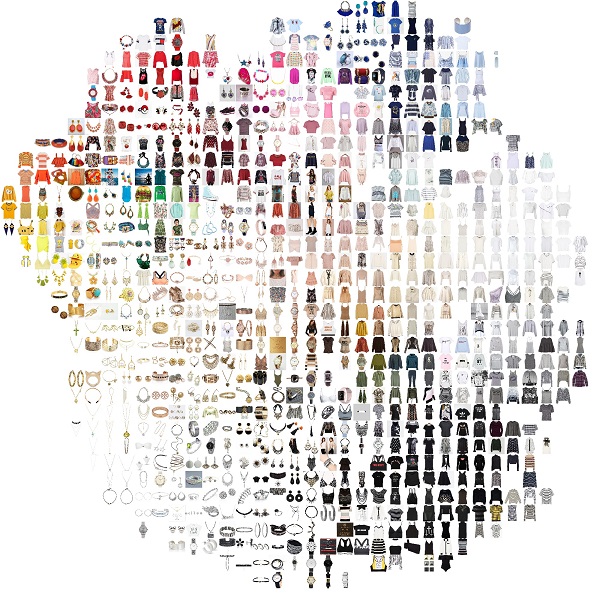}
\caption{\em \small t-SNE plot of the learned type-specific embedding space on our Polyvore dataset for tops and jewelery}
\end{figure}

\begin{figure}
\centering
\includegraphics[width=\textwidth]{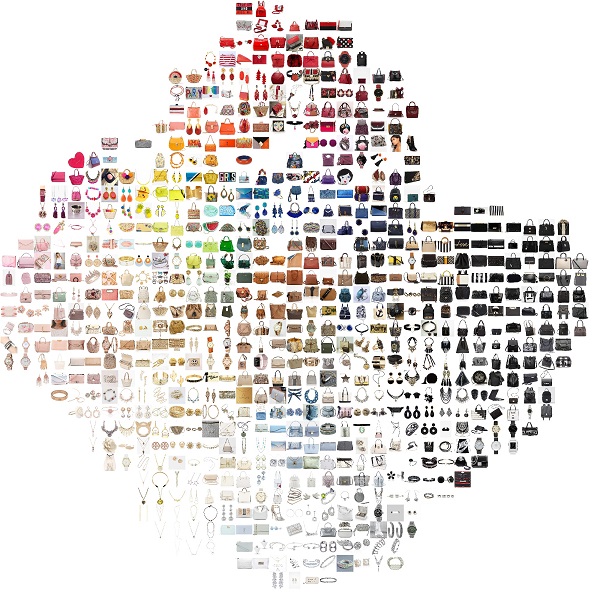}
\caption{\em \small t-SNE plot of the learned type-specific embedding space on our Polyvore dataset for bags and jewelery}
\end{figure}

\begin{figure}
\centering
\includegraphics[width=\textwidth]{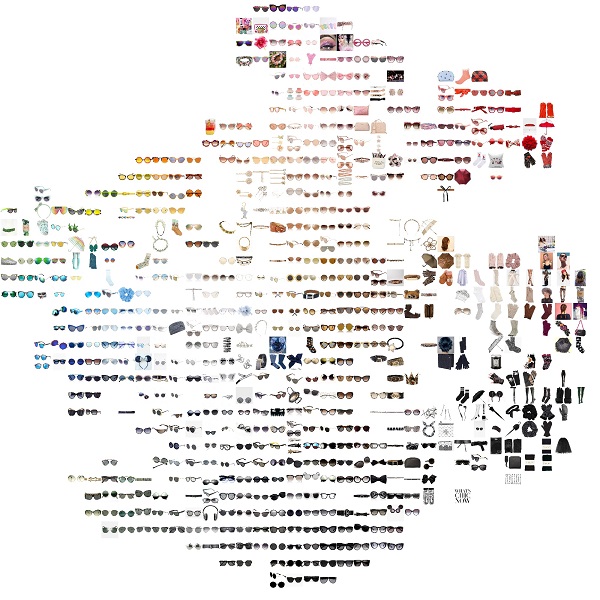}
\caption{\em \small t-SNE plot of the learned type-specific embedding space on our Polyvore dataset for sunglasses and accessories}
\end{figure}

\begin{figure}
\centering
\includegraphics[width=\textwidth]{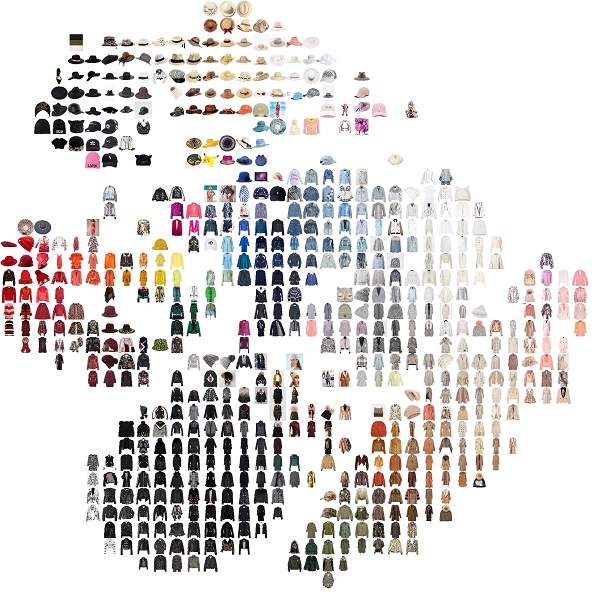}
\caption{\em \small t-SNE plot of the learned type-specific embedding space on our Polyvore dataset for hats and outerwear}
\end{figure}

\begin{figure}
\centering
\includegraphics[width=\textwidth]{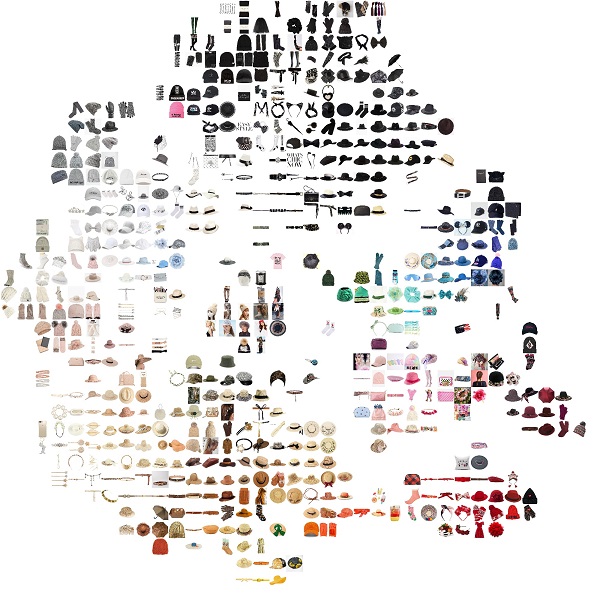}
\caption{\em \small t-SNE plot of the learned type-specific embedding space on our Polyvore dataset for hats and accessories}
\end{figure}

\begin{figure}
\centering
\includegraphics[width=\textwidth]{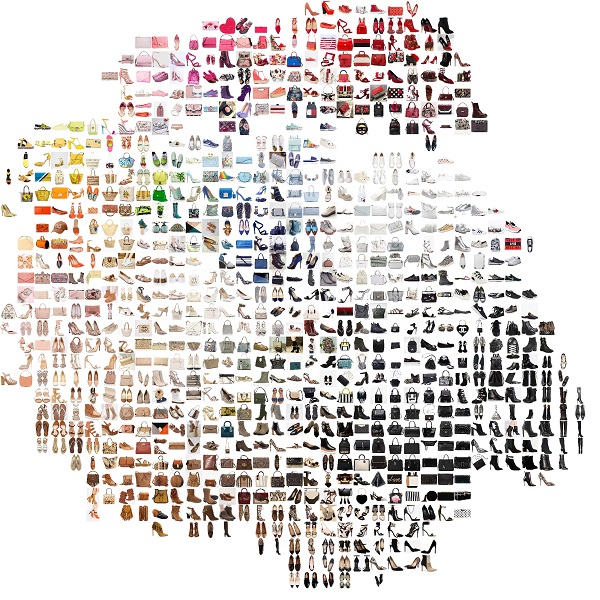}
\caption{\em \small t-SNE plot of the learned type-specific embedding space on our Polyvore dataset for bags and shoes}
\end{figure}

\begin{figure}
\centering
\includegraphics[width=\textwidth]{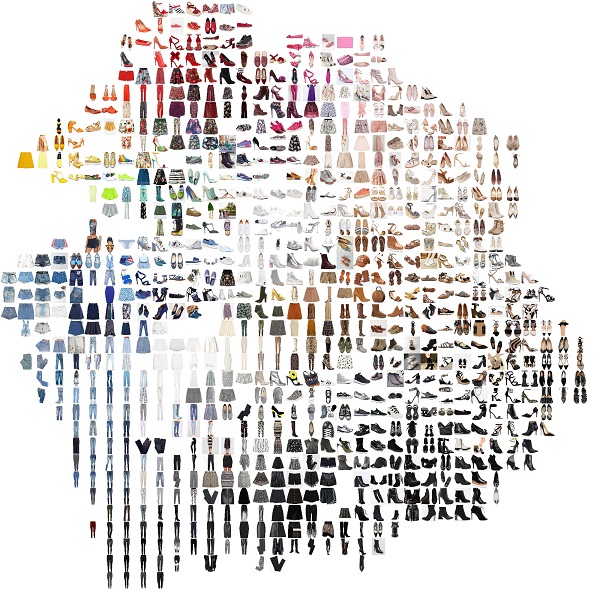}
\caption{\em \small t-SNE plot of the learned type-specific embedding space on our Polyvore dataset for bottoms and shoes}
\end{figure}